\documentclass[11pt]{article}
\usepackage[margin=3.5cm]{geometry}
\usepackage{amsmath,amssymb,amsthm}
\usepackage{hyperref}
\usepackage{titlesec}
\usepackage{graphicx}
\usepackage{subcaption}
\usepackage{booktabs}
\usepackage{xcolor}
\usepackage{longtable}
\usepackage{setspace}
\usepackage{multirow}
\theoremstyle{remark}
\newtheorem{remark}{Remark}

\titleformat{\subsection}[hang]{\normalfont\normalsize\bfseries}{\thesubsection}{1em}{}
\titleformat{\subsubsection}[hang]{\normalfont\normalsize\bfseries}{\thesubsubsection}{1em}{}

\title{\textbf{A Structured Nonparametric Framework for Nonlinear Accelerated Failure Time Models (KAN-AFT)}}

\author{
Mebin Jose$^{1}$ \and
Jisha Francis$^{1}$\thanks{Corresponding author: jishafrancis@vit.ac.in} \and
Sudheesh Kumar Kattumannil$^{2}$\thanks{sudheesh@isichennai.res.in}
}

\date{
$^{1}$Department of Mathematics, School of Advanced Sciences,\\
Vellore Institute of Technology, Vellore, India, 632014\\[6pt]
$^{2}$Applied Statistics Unit, Indian Statistical Institute, Chennai, India, 600029
}

\begin{document}

\maketitle

\begin{abstract}

Accelerated failure time (AFT) models provide a direct and interpretable time-scale description of covariate effects in lifetime data analysis, but classical formulations rely on linear predictors and are therefore limited in their ability to represent nonlinear relationships. Moreover, in heterogeneous clinical settings with complex covariate structures and varying censoring mechanisms, standard survival models such as the Cox proportional hazards model or AFT formulations may be inadequate due to restrictive structural assumptions.

We propose a structured nonparametric extension of the AFT framework in which the regression function governing log-survival time is an unknown smooth function represented through Kolmogorov--Arnold representations.
We formalize the nonlinear AFT estimand under independent right-censoring and show that the proposed function class strictly contains the classical linear AFT model as a special case. Estimation is carried out through a unified framework that accommodates several censoring-adjusted losses such as Buckley--James, inverse probability of censoring weight and transformation methods. Structural regularization and pruning promote parsimony, and symbolic approximation yields analytic representations of learned component functions.
Simulation studies show that the method recovers linear structure when appropriate and captures nonlinear effects when present. Applications to multiple clinical datasets demonstrate competitive predictive performance and transparent covariate-effect estimation.

\end{abstract}


\section{Introduction}
\label{Intro}
Lifetime data analysis  is a specialized statistical field concerned with modeling the time until the occurrence of a specific event, such as failure, relapse, or death. Unlike standard regression problems, survival data are characterized by censoring, most commonly right-censoring, where the exact event time is not always observed.  In many studies, it is only known that the event did not occur up to a certain observation time. Ignoring this partial information leads to biased estimates, making specialized survival models essential for reliable inference \cite{Klein2003survival}, \cite{Lee2003statistical},  \cite{Kleinbaum1996survival}, \cite{Wang2019MachineLearning}.

Beyond estimating baseline survival probabilities and modeling the temporal distribution of events, survival models aim to quantify the impact of specific covariates on event timing. In practice, this involves modeling the effect of covariates (individual features) such as age, creatinine level, ejection fraction, platelet count, serum markers, and sodium concentration on critical events (endpoints), such as heart failure-associated mortality. This enables tasks such as risk stratification, treatment comparison, and prediction of survival time under different conditions.

Over the past decades, two classes of models have played a central role in this field: the Cox Proportional Hazards (CoxPH) model and the Accelerated Failure Time (AFT) model. The CoxPH model \cite{Cox1972} is a semi-parametric framework that models the instantaneous risk of an event for individual $i$ as a function of the covariate vector $\mathbf{z}_i \in \mathbb{R}^{n_{\text{cov}}}$. The hazard function is defined as
\begin{equation*}
h_i(t \mid \mathbf{z}_i) =
h_0(t)\exp(\mathbf{z}_i^\top \boldsymbol{\beta}),
\quad t>0,
\end{equation*}
where $h_0(t)$ is an unspecified baseline hazard function and $\boldsymbol{\beta}$ is a vector of regression coefficients. The term $\mathbf{z}_i^\top \boldsymbol{\beta}$ represents a linear predictor that multiplicatively scales the baseline hazard. %The corresponding survival function is given by
%\begin{equation*}
%S(t \mid \mathbf{z}_i) =
%\left[S_0(t)\right]^{\exp(\mathbf{z}_i^\top \boldsymbol{\beta})},
%\quad t>0,
%\end{equation*}
%where
%\(
%S_0(t)=\exp\!\left(-\int_0^t h_0(u)\,du\right)
%\)
%denotes the baseline survival function. This formulation shows that covariates act multiplicatively on the baseline survival through a time-invariant effect.
While the CoxPH model avoids parametric assumptions on the hazard shape, it relies critically on the proportional hazards assumption, which requires covariate effects to remain constant over time. This assumption is frequently violated in practice, leading to unreliable inference and degraded predictive performance \cite{CoxKAN}.

In contrast to the hazard-based CoxPH model, the AFT model \cite{wei1992accelerated} directly relates covariates to the survival time itself, providing a time-centric interpretation of covariate effects \cite{kalbfleisch2002statistical}. Rather than modeling instantaneous risk, AFT models describe how covariates accelerate or decelerate the time to an event through a multiplicative time-scaling mechanism. The classical AFT model specifies the log survival time for the $i$th individual as
\begin{equation}
\log(T_i) =
\mathbf{z}_i^\top \boldsymbol{\beta}
+
\sigma \epsilon_i,
\label{Eq:AFT_Classical_Vector}
\end{equation}
where $T_i$ denotes the event time, $\mathbf{z}_i \in \mathbb{R}^{n_{\text{cov}}}$ is the covariate vector for individual $i$, $\boldsymbol{\beta}$ is a vector of regression coefficients, $\sigma>0$ is a scale parameter, and $\epsilon_i$ is a random error term with a specified distribution. %This formulation implies the survival function
%\begin{equation*}
%S(t \mid \mathbf{z}_i)
%=
%S_0\!\left(t\,\exp(-\mathbf{z}_i^\top \boldsymbol{\beta})\right),
%\end{equation*}
%where $S_0(t)$ denotes the baseline survival function for an individual with $\mathbf{z}_i=\mathbf{0}$.
Here, covariates act by rescaling time relative to the baseline distribution. The acceleration factor $\exp(\mathbf{z}_i^\top \boldsymbol{\beta})$ directly quantifies this scaling  effect, where values exceeding unity indicate a deceleration of the event process (prolonged survival), while values less than unity indicate an acceleration (shortened survival).

This intuitive time-scaling interpretation makes AFT models particularly appealing in applications where the timing of an event is of primary interest, such as lifespan analysis, recovery studies, and reliability modeling \cite{wei1992accelerated,hutton2002choice}. Moreover, AFT models remain applicable in settings where the proportional hazards assumption is violated \cite{Klein2003survival}. Despite these advantages, classical AFT models have notable limitations, which primarily arise from the restrictive assumption of a linear relationship between the logarithm of survival time and covariates \cite{kalbfleisch2002statistical}. This linearity assumption is often unrealistic in complex clinical, biological, and engineering systems, where covariate effects are frequently nonlinear and interacting.
% \textcolor{red}{they require specification of a parametric error distribution, which may lead to bias under misspecification,}

To address the limitations of classical CoxPH and AFT models, research has evolved along two distinct trajectories. Many early extensions emphasize interpretability through structured modeling, but rely on substantial manual specification of functional forms, basis expansions, or dependence structures, which limit scalability and flexibility in complex settings. Representative examples include Generalized Additive Models (GAMs) \cite{hastie1986generalized, hastie2017generalized, wood2025generalized}, structured additive regression models \cite{umlauf2015structured}, spline-based survival regression \cite{gray1992flexible, royston2002flexible}, and joint modeling frameworks \cite{tsiatis2004joint}. While these approaches retain transparency by construction, they often require careful basis selection, knot placement, or parametric assumptions, making them less suited to high-dimensional covariate spaces and complex nonlinear interactions without extensive manual tuning.

In parallel, flexible data-driven methods have been increasingly adopted as alternatives to rigid parametric survival models. Approaches such as random survival forests and boosting-based models accommodate nonlinear effects and interactions through adaptive partitioning and weighted likelihood formulations \cite{ishwaran2008random, chen2013gradient}. More recently, deep learning extensions replace linear predictors with highly flexible representations, including DeepSurv \cite{Katzman2018DeepSurv} and DeepHit \cite{lee2018deephit} for Cox-type models and DeepAFT \cite{Norman2024deepAFT} within the AFT framework. Although these methods often achieve strong predictive performance, they typically operate as black boxes, offering limited insight into covariate-specific effects and weakening interpretability in scientific and clinical applications \cite{CoxKAN}.

These limitations have motivated growing interest in regression frameworks that balance flexibility with structural transparency and proper handling of censoring. Structured nonparametric representations, in which regression functions are built from low-dimensional components such as univariate transformations, provide a promising direction. Kolmogorov–Arnold representations \cite{braun2009constructive, schmidt2021kolmogorov} supply a theoretical foundation by expressing multivariate functions through sums and compositions of univariate functions, enabling rich approximation while retaining interpretable structure. This perspective motivates a structured nonparametric extension of the AFT framework in which the effect of covariates on log-survival time is modeled as a smooth, structured function. The resulting formulation preserves the time-scaling interpretation of AFT models, relaxes restrictive linearity assumptions, and accommodates nonlinear effects and controlled interactions, while supporting robust estimation under right-censoring through established censoring-adjusted loss constructions.

We evaluate the proposed framework through simulation studies and analyses of multiple real-world clinical survival datasets. The simulations consider both linear and nonlinear settings to examine whether the method reduces to simple parametric structure when appropriate and adapts to nonlinear effects when present. The real-data applications include lung cancer, breast cancer, liver disease, and heart failure, spanning a range of censoring levels and covariate complexity. Across these studies, the framework achieves predictive performance that is competitive with or superior to classical parametric AFT models and deep learning–based approaches, while yielding transparent and clinically meaningful covariate-effect representations.

The remainder of the paper is organized as follows. Section~\ref{Section:GeneralisedAFT_Methodology} introduces a structured nonparametric formulation of AFT models and defines the associated statistical estimand. Section~\ref{section:KAN-AFT} presents a Kolmogorov-Arnold network based realization of this framework, detailing the model architecture, estimation procedures, regularization and pruning strategies, and methods for handling right-censored data. %; this section also establishes basic theoretical properties of the proposed estimator, including its relationship to the classical linear AFT model and consistency under mild regularity conditions.
Section~\ref{Section:SimulationStudy} investigates finite-sample and empirical performance through simulation studies and applications to multiple real-world survival datasets, with emphasis on predictive accuracy and interpretability. Section~\ref{Section:Conclusion} concludes with a discussion and directions for future research.

%%%%%%%%%%%%%%%%%%%%%%%%%%%%%%%%%%%%%%%%%%%%%%%%%%%%%%%%%%%%%%%%%%%%%%%%%%%%%
\section{A Structured Nonparametric Framework for AFT Models}
\label{Section:GeneralisedAFT_Methodology}

\subsection{A Generalized AFT Framework}
\label{subsection2_1}

The classical AFT model in Equation~\eqref{Eq:AFT_Classical_Vector} provides a transparent description of how covariates influence event timing through a linear predictor. While attractive for its interpretability, this formulation assumes that covariate effects on the logarithm of survival time are linear and additive. Empirical evidence in many applications, however, suggests that such effects may be nonlinear or vary across the covariate domain.

To retain the defining time-scale interpretation of AFT models while allowing greater flexibility, we consider a  generalized formulation in which the linear predictor is replaced by an unknown regression function $\eta(\cdot)$:
\begin{equation}
\log(T_i) =
\eta(\mathbf{z}_i) +
\sigma \epsilon_i,
\label{Eq:AFT_Nonlinear}
\end{equation}
where $\mathbf{z}_i \in \mathbb{R}^{n_{\text{cov}}}$ is the covariate vector, $\sigma>0$ is a scale parameter, and $\epsilon_i$ is a mean-zero error term independent of $\mathbf{z}_i$.

%This formulation implies the survival function
%\[
%S(t \mid \mathbf{z}_i)
%=
%S_0\!\left(t \exp\{-\eta(\mathbf{z}_i)\}\right),
%\]
%so that covariates act multiplicatively on the time scale through the acceleration factor $\exp\{\eta(\mathbf{z}_i)\}$. The familiar interpretation of covariate effects as accelerating or decelerating the event process is therefore preserved, irrespective of the functional form of $\eta(\cdot)$.

This generalized AFT formulation defines the statistical estimand, the covariate-dependent acceleration function, without specifying how it is represented or estimated. This separation provides a flexible foundation for modeling complex covariate effects while maintaining the core interpretive structure that motivates AFT models.

\subsection{Structured Nonparametric AFT Regression}
\label{subsection2_2}

Under the generalized AFT formulation in Equation~\eqref{Eq:AFT_Nonlinear}, the regression function $\eta(\cdot)$ governs the conditional mean of $\log(T)$ given the covariates and is treated as an unknown population-level function. To balance flexibility with interpretability, we restrict $\eta(\cdot)$ to a structured nonparametric function class in which nonlinear effects are localized to univariate transformations of individual covariates, and multivariate dependence arises only through explicit and shallow aggregation. This structure permits nonlinear main effects and controlled interactions while preserving clear marginal interpretations and encouraging parsimony through pruning when interaction effects are unsupported by the data.

This formulation cleanly separates the statistical model from its numerical realization. The function $\eta(\cdot)$ defines the nonparametric AFT estimand, while estimation is carried out using a structured sieve approximation whose complexity is regulated through regularization and pruning. As a result, the framework relaxes the linearity assumption of classical AFT models while preserving the defining time-scaling interpretation via the acceleration factor $\exp{(\eta(\mathbf{z}))}$.

%A concrete implementation of this structured estimator is introduced in the following section.
%%%%%%%%%%%%%%%%%%%%%%%%%%%%%%%%%%%%%%%%%%%%%%%%%%%%%%%%%%%%%%%%%%%%%%%%
%%%%%%%%%%%%%%%%%%%%%%%%%%%%%%%%%%%%%%%%%%%%%%%%%%%%%%%%%%%%%%%%%%%%%%%%
\section{Kolmogorov-Arnold Representations for Structured AFT Models}
\label{section:KAN-AFT}
This section presents a practical structured nonparametric representation of the AFT regression function $\eta(\cdot)$ based on Kolmogorov-Arnold representations, yielding an implementable estimator that accommodates nonlinear covariate effects while preserving interpretability.

%%%%%%%%%%%%%
\subsection{Kolmogorov-Arnold Networks}
\label{subsection3_1}

Kolmogorov-Arnold representations characterize multivariate functions through sums and compositions of univariate functions. In particular, the Kolmogorov-Arnold representation theorem states that any continuous function $f:\mathbb{R}^{n_0}\to\mathbb{R}$ can be approximated arbitrarily well by a finite sum of compositions of univariate continuous functions. This result motivates a structured nonparametric function class in which nonlinearity is localized to univariate transformations, while multivariate dependence arises only through explicit aggregation. The resulting structure provides a natural compromise between modeling flexibility and interpretability.

Kolmogorov-Arnold Networks (KANs) offer a concrete parameterization of this representation by associating each univariate component with a flexible function estimator. Below, we outline the KAN construction and contrast it with standard multilayer perceptrons (MLPs) to highlight the key structural differences.

To facilitate a transparent comparison and to align with the Kolmogorov–Arnold representation, both MLPs and KANs are considered under the same single hidden-layer architecture $[n_0, n_h, 1]$, comprising input layer ($l=0$) with $n_0$ nodes, hidden layer ($l=1$) with $n_h$ nodes, and output layer ($l=2$) with a single node producing a scalar response $\hat{y} \in \mathbb{R}$. The corresponding input vector is $\mathbf{x} = (x_1, x_2, \dots, x_{n_0})^\top \in \mathbb{R}^{n_0}$.

An MLP approximates an unknown multivariate function by composing affine linear transformations with fixed nonlinear activation functions. At each node, the input variables are first combined through a weighted linear sum and subsequently passed through a predefined nonlinear activation function.  The single hidden layer $[n_0, n_h, 1]$ MLP network defines the mapping
\begin{equation*}
\label{eq:mlp_output}
\hat{y}
=
\sigma_{\mathrm{out}}
\left(
\sum_{k=1}^{n_h}
w_{1,k,1}
\,
\sigma_{\mathrm{k}}
\left(
\sum_{j=1}^{n_0}
w_{0,j,k} x_j
+
b_{1,k}
\right)
+
b_{2}
\right),
\end{equation*}
where $w_{l,q,r} \in \mathbb{R}$ denotes the scalar weight connecting node $q$ in layer $l$ to node $r$ in layer $l+1$, and $b_{l,q} \in \mathbb{R}$ denotes the bias associated with node $q$ in layer $l$. Since the output layer contains a single node, $b_2$ is a scalar bias. The functions $\sigma_{\mathrm{k}}(\cdot)$ and $\sigma_{\mathrm{out}}(\cdot)$ denote the fixed nonlinear activation function applied a priori at each hidden node and the output activation function, respectively.

%During training, learning is restricted to scalar weights and biases, while the nonlinear activation functions remain fixed; thus, all nonlinearity is imposed a priori, and the resulting representation of input contributions is implicit.

KANs \cite{KANOriginal} draw theoretical inspiration from the Kolmogorov-Arnold representation theorem \cite{braun2009constructive}, which states that any continuous multivariate function can be expressed as a finite sum of compositions of univariate continuous functions. The single hidden layer $[n_0, n_h, 1]$ KAN network defines the mapping
\begin{equation*}
\label{eq:kan_output}
\hat{y}
=
\text{KAN}(\mathbf{x})
=
\sum_{k=1}^{n_h}
\phi_{1,k,1}
\left(
\sum_{j=1}^{n_0}
\phi_{0,j,k}(x_j)
\right),
\end{equation*}
where $\phi_{l,q,r}(\cdot)$ denotes a learnable univariate function associated with the edge connecting node $q$ in layer $l$ to node $r$ in layer $l+1$. The inner summation aggregates the transformed contributions of each input variable $x_j$ to the $k$-th hidden node, while the outer summation combines the transformed hidden-layer outputs into the final scalar prediction.

In a KAN, all nonlinear behavior is explicitly captured by the univariate edge functions $\phi_{l,q,r}(\cdot)$, while nodes themselves perform only linear aggregation. This explicit separation of nonlinearity and aggregation fundamentally distinguishes KANs from conventional MLPs.

For compactness, the single hidden layer KAN can also be expressed as a composition of layer operators,
\begin{equation*}
\hat{y}
=
\boldsymbol{\Phi}_1
\circ
\boldsymbol{\Phi}_0
(\mathbf{x}),
\end{equation*}
where $\boldsymbol{\Phi}_0 : \mathbb{R}^{n_0} \rightarrow \mathbb{R}^{n_h}$ and $\boldsymbol{\Phi}_1 : \mathbb{R}^{n_h} \rightarrow \mathbb{R}$ denote collections of univariate functions defining the layer mappings whose entries are given by $\phi_{l,q,r}(\cdot)$.

Each edge function $\phi_{l,q,r}(\cdot)$ is parameterized as a combination of a fixed base function $b(x)$  and a flexible B-spline expansion:
\begin{equation*}
\phi(x) = w_b\, b(x) + w_s \sum_{m=0}^{G+d-1} c_m B_{m,d}(x),
\end{equation*}
where $w_b$ and $w_s$ are trainable scaling weights, $c_m$ are spline coefficients, and $B_{m,d}(x)$ denotes B-spline basis functions of degree $d$ defined over $G$ grid intervals. Each edge function is equipped with its own set of spline coefficients, enabling input-specific nonlinear transformations and supporting fine-grained interpretability of individual effects.

\subsection{Estimation Framework for Structured Nonlinear AFT Models}
\label{subsection3_2}

In the structured nonlinear AFT framework of Section~\ref{Section:GeneralisedAFT_Methodology}, the regression function
\(
\eta : \mathbb{R}^{n_{\mathrm{cov}}} \to \mathbb{R}
\)
is an unknown multivariate function of the covariate vector $\mathbf{z}$, restricted to a structured nonparametric function class defined through compositions of univariate functions. For estimation, this function is approximated using a Kolmogorov-Arnold representation. Without loss of generality, the KAN input is taken to coincide with the covariate vector, so that $\mathbf{x}_i \equiv \mathbf{z}_i \in \mathbb{R}^{n_0}$ with $n_0 = n_{\mathrm{cov}}$. The nonlinear predictor is therefore parameterized as

\begin{equation*}
\label{eq:KAN_NonlinearAFT}
\eta(\mathbf{z}_i) = \mathrm{KAN}(\mathbf{z}_i).
\end{equation*}

Let $\{(T_i, C_i, \mathbf{z}_i)\}_{i=1}^n$ denote an independent sample of size $n$, where $T_i$ is the latent event time for individual $i$, $C_i$ is the corresponding censoring time, and $\mathbf{z}_i \in \mathbb{R}^{n_{\mathrm{cov}}}$ is the associated covariate vector. The observed data consist of $(\tilde{T}_i, \delta_i, \mathbf{z}_i)$, where $\tilde{T}_i = \min(T_i, C_i)$ and $\delta_i = \mathbb{I}(T_i \le C_i)$. Define $Y_i = \log(\tilde{T}_i)$.

In the absence of censoring, estimation of the nonlinear AFT regression function $\eta(\cdot)$ may be formulated through the squared-error loss function
\begin{equation}
\label{eq:aft_loss}
L_{\mathrm{AFT}}
=
\frac{1}{n}\sum_{i=1}^{n}
\left(
Y_i - \mathrm{KAN}(\mathbf{z}_i)
\right)^2,
\end{equation}
which corresponds to estimating the regression function $\mathbb{E}[\log(T_i)\mid \mathbf{z}_i]$.

\subsubsection{Estimation under Right-Censoring}
\label{subsubsection3_2_1a}

Under right-censoring, the idealized objective discussed in Equation \eqref{eq:aft_loss} is no longer directly applicable, as $\log(T_i)$ is unobserved when $\delta_i=0$. To accommodate censoring, we replace $L_{\mathrm{AFT}}$ by a censoring-aware loss, denoted $L_{\mathrm{survival}}$, whose form depends on the chosen censoring-handling strategy. The resulting estimation problem is defined through the regularized objective
\begin{equation}
\label{eq:total_loss}
L_{\mathrm{Total}}
=
L_{\mathrm{survival}}
+
L_R,
\end{equation}
where $L_{\mathrm{survival}}$ denotes a censoring-adjusted AFT loss and $L_R$ enforces structural regularization of the KAN parameterization.

The objective function $L_{\mathrm{Total}}$ combines a censoring-aware regression loss with structural regularization and is optimized using gradient-based methods. This differentiable loss provides a suitable surrogate for parameter estimation under right-censoring. Model selection and structural decisions, however, are guided by the concordance index (C-index), which evaluates the concordance of predicted survival times in the presence of censoring. %The C-index is not optimized directly; instead, it is used as an external criterion to assess predictive ranking performance and to inform pruning decisions and architectural selection.

We consider three established censoring-handling strategies adapted from the DeepAFT framework: iterative Buckley--James imputation, inverse probability of censoring weights (IPCW), and a censoring-adjusted transformation. These strategies represent complementary trade-offs. Buckley--James employs iterative residual-based imputation and is robust under heavy censoring at increased computational cost. IPCW yields a non-iterative, weighted objective that is computationally efficient but depends on accurate estimation of the censoring distribution. The transformation-based approach embeds censoring adjustment directly into the regression target, simplifying optimization while increasing sensitivity under substantial censoring. Detailed algorithms for the censoring-handling strategies are provided in \ref{Append_Censoring}.

\subsubsection{Regularization and Structural Sparsity}
\label{subsubsection3_2_1b}

The regularization term $L_R$ in Equation \eqref{eq:total_loss} promotes parsimony and interpretability of the KAN parameterization and is defined as
\begin{equation*}
\label{eq:regularization}
L_R
=
\sum_{l=0}^{1}
\left(
\|\boldsymbol{\Phi}_l\|_1
+
\lambda_1 S(\boldsymbol{\Phi}_l)
+
\lambda_2 \|\mathbf{C}_l\|_1
\right),
\end{equation*}
where $\boldsymbol{\Phi}_l$ denotes the matrix of univariate edge functions in layer $l$ and $\mathbf{C}_l$ the corresponding spline coefficient matrix. The quantity $\|\boldsymbol{\Phi}_l\|_1$ denotes the sum of the $L_1$ norms of all edge functions in layer $l$ and induces edge-level sparsity by shrinking negligible functions toward zero. The entropy  $S(\boldsymbol{\Phi}_l)$ defined by
\[
S(\boldsymbol{\Phi}_l)
=
-
\sum_{j,k}
\frac{\|\phi_{l,j,k}\|_1}{\|\boldsymbol{\Phi}_l\|_1}
\log
\left(
\frac{\|\phi_{l,j,k}\|_1}{\|\boldsymbol{\Phi}_l\|_1}
\right)
\]
 constructed from the relative $L_1$ norms of the univariate edge functions within layer $l$ encourages modular functional pathways, while the penalty $\|\mathbf{C}_l\|_1$ favors smooth, low-complexity spline representations.

Consistent with the principle of parsimony, estimation is carried out in a staged manner. The procedure begins with a purely additive KAN representation, corresponding to a network without a hidden layer, in which the regression function is expressed as a sum of univariate transformations of the covariates. This additive specification provides a highly interpretable baseline and represents the simplest model within the proposed structured nonparametric class. In this case, the nonlinear AFT regression function takes the form
\[
\eta(\mathbf{z})
=
\sum_{j=1}^{n_{\mathrm{cov}}}
\tilde{\phi}_{0,j,1}(z_j).
\]

Only when the additive representation proves inadequate to capture the observed dependence structure is the model extended to a KAN architecture with a single hidden layer. In this setting, $l=0$ denotes input--hidden edges and $l=1$ denotes hidden--output edges. This shallow architecture is sufficient for universal approximation under the Kolmogorov-Arnold representation theorem, while remaining compatible with interpretability and identifiability considerations and avoiding the complexity associated with deeper functional compositions. The resulting regression function is given by
\begin{equation} \label{eqn:1}
 \eta(\mathbf{z})
=
\sum_{k=1}^{n_h}
\phi_{1,k,1}
\left(
\sum_{j=1}^{n_{\mathrm{cov}}}
\phi_{0,j,k}(z_j)
\right).
\end{equation}

Following estimation with sparsity-inducing regularization, further simplification is performed through explicit node-level pruning. For the $k$th hidden-layer node ($l=1$), define the incoming and outgoing importance scores
\[
I_{1,k}
=
\max_{j}\bigl\|\phi_{0,j,k}\bigr\|_1,
\qquad
O_{1,k}
=
\bigl\|\phi_{1,k,1}\bigr\|_1,
\]
where $\phi_{0,j,k}$ and $\phi_{1,k,1}$ denote the univariate edge functions on the input--hidden and hidden--output connections, respectively. Hidden nodes satisfying
\(
\min\!\left(I_{1,k},\, O_{1,k}\right) < \theta,
\)
a threshold, are pruned.

The following proposition establishes that the proposed regression function in Equation \eqref{eqn:1} strictly generalizes the classical linear AFT model, which arises as a special case under simple linearity constraints on the component functions.
\begin{remark}[Linear AFT as a Special Case]
Suppose that
\(
\phi_{0,j,k}(z_j) = a_{j,k} z_j,
\quad
\phi_{1,k,1}(u) = b_k u.
\)
Then the proposed regression function reduces to a linear function of $z$, and the resulting model coincides with the classical AFT formulation.
\end{remark}

\begin{proof}
Under the stated assumptions,
\[
\eta(z)
= \sum_{k=1}^{n_h} b_k \sum_{j=1}^{n_{\mathrm{cov}}} a_{j,k}z_j
=
\sum_{j=1}^{{n_{\mathrm{cov}}}}
\Bigl(
\sum_{k=1}^{n_h} b_k a_{j,k}
\Bigr) z_j,
\]
Defining \(\beta_j= \sum_{k=1}^{n_h} b_k a_{j,k}\),  we obtain
\[
\eta(z)
=
\sum_{j=1}^{n_{\mathrm{cov}}}
\beta_j z_j
=
\beta^\top z.
\]
Substituting into
\(
\log T = \eta(Z) + \varepsilon
\)
yields the classical linear AFT model.
\end{proof}
\begin{remark}
Remark~1 implies that the proposed framework is backward compatible with classical AFT modeling: when the true relationship is linear, the structured representation can recover the standard linear AFT form without imposing nonlinear effects.
\end{remark}
%%%%%%%%%%%%%%%%%%%%%%%%%%%%%%%%%%
\subsubsection{Symbolic Regression and Interpretability}
\label{subsection3_2_2}

Following sparsity-induced pruning, the remaining spline-based univariate component functions are approximated using symbolic regression,
\(
\phi_{l,j,k}(z) \approx \tilde{\phi}_{l,j,k}(z),
\)
with the goal of obtaining explicit analytical representations.

Following \cite{KANOriginal}, each learned component function is approximated by fitting an affine transformation of candidate elementary functions drawn from a predefined library.

The symbolic approximation proceeds in stages. Simple functions, such as linear or low-degree polynomial functions, are considered first to favor parsimony. If no candidate achieves an adequate \(R^2\) threshold, the procedure is extended to a broader library of elementary functions, including exponential, logarithmic, and trigonometric forms. Only when these candidates fail to provide a satisfactory fit is a more flexible evolutionary symbolic regression search invoked using the PySR library \cite{cranmer2023interpretable}.

In the AFT context, the resulting symbolic expressions directly characterize how individual covariates accelerate or decelerate survival time. For example, \(\tilde{\phi}(z)=\exp(z)\) corresponds to exponential time acceleration, while \(\tilde{\phi}(z)=\log(1+z^2)\) reflects diminishing marginal effects. This procedure preserves predictive fidelity while yielding transparent and interpretable covariate effects.

Hereafter, we refer to the proposed estimator as KAN-AFT,  highlighting that the nonparametric AFT regression function is represented  using a Kolmogorov-Arnold sieve. This terminology reflects the chosen  functional approximation and does not alter the underlying AFT model or  its statistical interpretation.
%%%%%%%%%%%%%%%%%%%%%%%%%%%%%%%%%%%%%%%%%%%%%%%%%%%%%%%%

%%%%%%%%%%%%%%%%%%%%%%%%%%%%%%%%%%%%%%%%%%%%%%%%%%%%%%%%%%%%%%%%%%
\section{Simulation Study and Data Analysis}
\label{Section:SimulationStudy}
\subsection{Simulation design and evaluation criteria}

Estimation under the proposed KAN-AFT framework follows a structured three-stage procedure, consistent with the methodology developed in Sections~\ref{Section:GeneralisedAFT_Methodology} and~\ref{section:KAN-AFT}. First, the regression function is estimated by minimizing the regularized censoring-aware loss in Equation~\eqref{eq:total_loss} using the L-BFGS optimizer with a maximum of $50$ iterations. The limited-memory quasi-Newton scheme is adopted for its numerical stability and its ability to produce smooth functional estimates, which is advantageous for subsequent symbolic approximation. Early stopping is employed to prevent overfitting; training terminates when the validation loss fails to improve by at least $10^{-3}$ for a prespecified number of consecutive iterations (patience).

Second, structural simplification is performed via explicit pruning. Following estimation, univariate component functions with $L_1$ norm below the threshold $\theta=10^{-2}$ are removed together with their associated connections. This step reduces estimator complexity, enhances interpretability, and enforces the principle of parsimony described in Section~\ref{subsubsection3_2_1b}. The hidden-layer width is initialized in the range \( n_0 \leq n_h \leq 2n_0+1 \), where \( n_0 \) denotes the number of covariates, and is subsequently reduced through pruning to adapt model complexity to the data.

Third, the remaining univariate component functions are approximated via symbolic regression to obtain explicit analytical expressions. Following \cite{KANOriginal}, each component is approximated by an affine transformation of candidate elementary functions from a predefined \texttt{auto\_symbolic} library,
\(
\tilde{\phi}(z)=c\,f(a z+b)+g,
\)
with affine parameters $(a,b,c,g)$ estimated by least squares. Approximation quality is assessed using the coefficient of determination $R^2$. If no candidate attains the prespecified accuracy threshold, a broader symbolic search is conducted using the \texttt{PySR} library \cite{cranmer2023interpretable}.

The KAN-AFT model is evaluated under three censoring-handling strategies: Buckley--James imputation, inverse probability of censoring weighting (IPCW), and a censoring-adjusted transformation. For all experiments, data are randomly partitioned into training and test sets in an $80{:}20$ ratio. Tuning parameters controlling model flexibility and regularization are selected using random search \cite{bergstra2012random} with five-fold cross-validation, using the average concordance index (C-index) as the selection criterion. The tuning is performed over spline degree \(d\in\{3,5\}\), grid size \(G\in\{5,10,15\}\), base function \(b(\cdot)\in\{x,\mathrm{silu}(x)\}\), regularization weight (decay) \(\lambda_2\in\{0.01,0.02,0.03,0.04,0.05,0.06\}\) with \(\lambda_1=0.01\,\lambda_2\), patience $\in\{5,7,10,15\}$, and pruning threshold \(\theta=10^{-2}\).

Model performance is assessed using the concordance index (C-index) and the mean squared error (MSE) of $\log(T)$. Comparative results are reported for classical parametric AFT models, DeepAFT, and the proposed KAN-AFT under each censoring strategy. Detailed hyperparameter settings for all real-data applications are provided in \ref{append_3}.

%The empirical evaluation includes both synthetic datasets with linear and nonlinear data-generating mechanisms and multiple real-world survival datasets, enabling a comprehensive assessment of predictive accuracy, robustness to censoring, and structural recovery of covariate effects.

%%%%%%%%%%%%%%%%%%%%%%%%%%%%%%%%%%%%%%%%%%%%%%%%%%%%%%%%%%
\subsection{Linear synthetic data}

The linear synthetic dataset consists of $n=1000$ independent observations, each comprising an observed time, an event indicator, and three covariates. The covariate vector $\mathbf{z}=(z_1,z_2,z_3)^\top$ has independent components drawn from the standard normal distribution.

Latent event times are generated according to a linear AFT model. Specifically, the logarithm of the latent event time $T'$ is defined as
\begin{equation}
\label{Eq_Simulation_linear}
\log(T') = 0.5\, z_1 - 0.3\, z_2 + z_3 + \epsilon,
\end{equation}
where $\epsilon \sim \mathcal{N}(0,0.5)$ independently across observations, consistent with the classical AFT formulation in Equation~\eqref{Eq:AFT_Classical_Vector}.

Right-censoring is introduced by generating censoring times $C \sim \mathrm{Exp}(\lambda=10)$ independently of $T'$. The observed data consist of $(\tilde T,\delta)$, where $\tilde T=\min(T',C)$ and $\delta=\mathbb{I}(T'\le C)$.

\begin{figure}[h!tbp]
    \centering
    \includegraphics[width=0.75\linewidth]{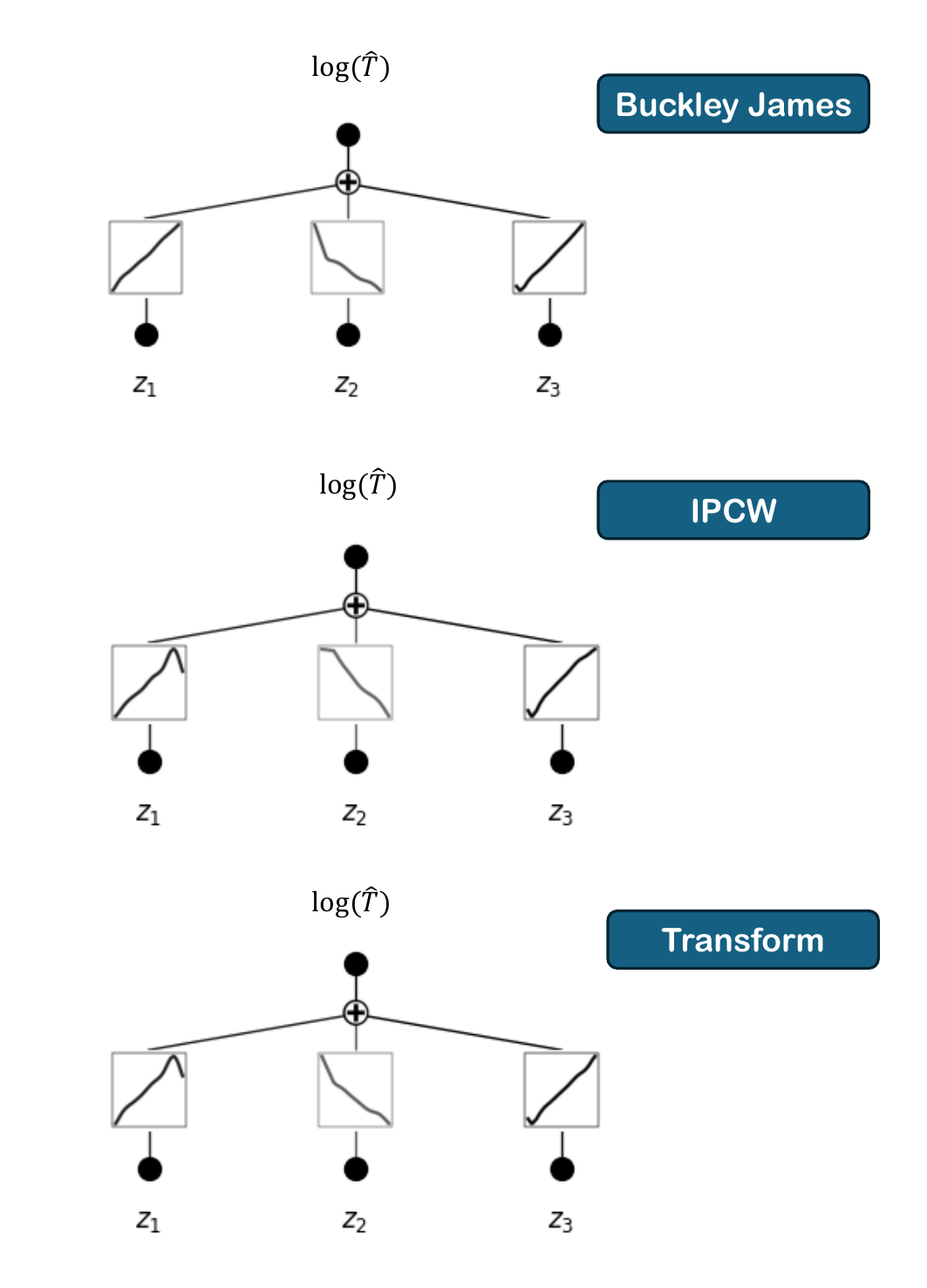}
    \caption{Estimated univariate component functions in the KAN-AFT model for linear synthetic data.}
    \label{fig:1}
\end{figure}

Figure~\ref{fig:1} presents the estimated univariate component functions obtained under the Buckley--James, IPCW, and transformation-based censoring-handling strategies. After symbolic simplification, the corresponding regression functions for $\log(\hat{T})$ are provided in Equations~\eqref{eq:lin_bj}--\eqref{eq:lin_trans}.
\vspace{-.4cm}
\begin{subequations}
\label{eq:lin_symbolic_models}
\begin{align}
\log(\hat{T})
&=
0.5113\, z_1 - 0.3081\, z_2 + 0.9483\, z_3 - 0.0556,
\label{eq:lin_bj}\\[4pt]
\log(\hat{T})
&=
0.4043\, z_1 - 0.2086\, z_2 + 0.7463\, z_3 - 0.0842,
\label{eq:lin_ipcw}\\[4pt]
\log(\hat{T})
&=
0.4556\, z_1 - 0.2529\, z_2 + 0.8480\, z_3 - 0.0208.
\label{eq:lin_trans}
\end{align}
\end{subequations}

The estimates in Equation~\eqref{eq:lin_symbolic_models} closely approximate the true regression function in Equation~\eqref{Eq_Simulation_linear}, up to sampling variability and the effects of censoring.

Across all censoring strategies, the estimated univariate component functions are approximately linear, and all additional components are eliminated through pruning. Consequently, the fitted KAN-AFT model reduces to a purely additive linear representation. This behavior demonstrates that the proposed framework does not introduce spurious nonlinearity when the underlying relationship is linear, and instead recovers a parsimonious model aligned with the true data-generating process.

Table~\ref{tab:model_comparison1} summarizes predictive performance. KAN-AFT achieves C-index values comparable to correctly specified parametric AFT models and consistently higher than those obtained by DeepAFT. In addition, KAN-AFT yields substantially lower MSE, reflecting accurate estimation of the conditional mean $\mathbb{E}[\log(T)\mid \mathbf{z}]$.

\begin{table}[h!tbp]
\centering
\caption{Comparison of AFT estimators on linear synthetic data}
\label{tab:model_comparison1}
\begin{tabular}{@{}l cccc@{}}
\toprule
Model & \multicolumn{2}{c}{C-index} & \multicolumn{2}{c}{MSE}\\
\cmidrule(r){2-3}\cmidrule(l){4-5}
& Train & Test & Train & Test\\
\midrule
AFT-Weibull & 0.8684 & 0.8724 & 1.1223 & 0.9754\\
AFT-Log-normal & 0.8686 & 0.8741 & 1.1389 & 0.8755\\
AFT-Log-logistic & 0.8684 & 0.8741 & 1.1376 & 0.8724\\
\midrule
DeepAFT-BuckleyJames &  0.8241  & 0.7835 & 4.3393 & 3.2239\\
DeepAFT-IPCW & 0.8070  & 0.8018 & 2.2439 & 1.1326\\
DeepAFT-Transform & 0.8158  & 0.8107  & 3.9888 & 2.2821\\
\midrule
KAN-AFT-BuckleyJames & \textbf{0.8690} & \textbf{0.8750} & \textbf{0.2500} & \textbf{0.2879}\\
KAN-AFT-IPCW & 0.8641 & 0.8691 & 0.3333 & 0.3554\\
KAN-AFT-Transform & 0.8675 & 0.8727 & 0.2616 & 0.3072\\
\bottomrule
\end{tabular}
\end{table}

%%%%%%%%%%%%%%%%%%%%%%%%%%%%%%%%%%%%%%%%%%%%%%%%%%%%%%%%%%
\subsection{Nonlinear synthetic data}

The nonlinear synthetic dataset consists of $n=1000$ independent observations, each comprising an observed time, an event indicator, and three covariates. The covariate vector $\mathbf{z}=(z_1,z_2,z_3)^\top$ has independent components drawn from the standard normal distribution.

Latent event times are generated from a nonlinear accelerated failure time model. Specifically,
\begin{equation}
\label{Eq_Simulation_Nonlinear}
\log(T')
=
0.5\, z_1^2
+
0.3\, \exp(z_2)
+
0.8\, \sin(z_3)
+
\epsilon,
\end{equation}
where $\epsilon \sim \mathcal{N}(0,0.5)$ independently across observations. This specification induces heterogeneous nonlinear covariate effects while preserving the multiplicative time-scaling interpretation of the AFT framework.

Right-censoring is introduced by generating censoring times $C \sim \mathrm{Exp}(\lambda=10)$ independently of $T'$. The observed data consist of $(\tilde T,\delta)$, where $\tilde T=\min(T',C)$ and $\delta=\mathbb{I}(T'\le C)$.

\begin{figure}[h!tbp]
    \centering
    \includegraphics[width=0.75\linewidth]{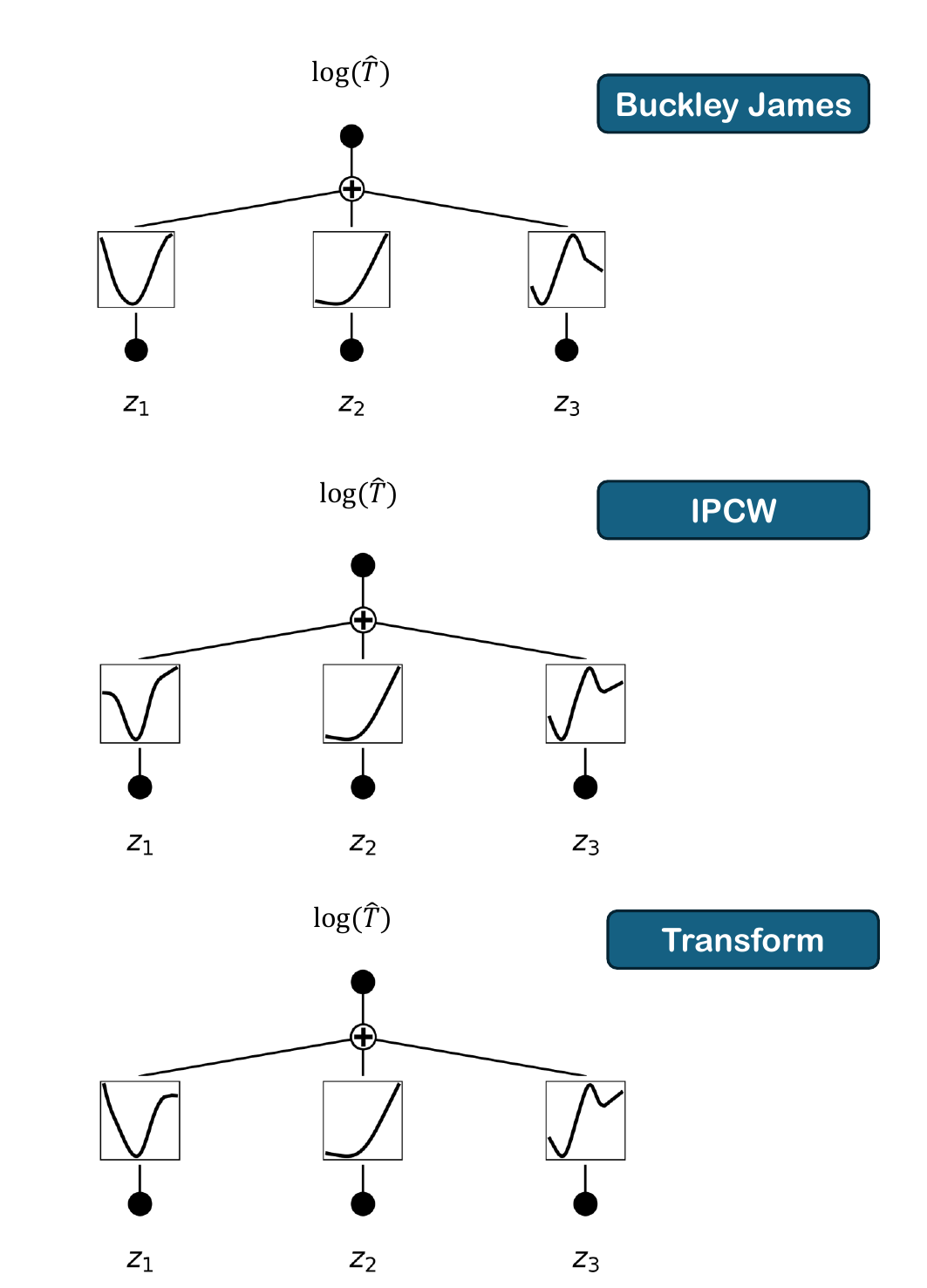}
    \caption{Estimated univariate component functions in the KAN-AFT model for nonlinear synthetic data.}
    \label{fig:2}
\end{figure}

Figure~\ref{fig:2} shows the estimated univariate component functions obtained from the KAN-AFT model after regularization and pruning. In contrast to the linear case, the recovered components exhibit clear nonlinear structure, capturing the quadratic, exponential, and periodic effects specified in Equation~\eqref{Eq_Simulation_Nonlinear}. The corresponding symbolic approximations derived under the Buckley--James, IPCW, and transformation-based KAN-AFT fits are reported in Equations~\eqref{eq:nl_bj}--\eqref{eq:nl_trans}.
\vspace{-1cm}

\begin{subequations}
\label{eq:nl_symbolic_models}
\begin{align}
\log(\hat{T}) &=
0.016(0.200 - 5.386 z_1)^2
+ 0.778 \exp(0.510 z_2)
+ 0.840 \sin(1.022 z_3 + 6.383)
- 1.380,
\label{eq:nl_bj}\\[4pt]
\log(\hat{T}) &=
0.003(-8.164 z_1 - 0.621)^2
+ 0.375 \exp(0.510 z_2)+ 0.471 \sin(1.223 z_3 + 3.396) - 0.804,
\label{eq:nl_ipcw}\\[4pt]
\log(\hat{T}) &=
0.024(0.032 - 3.511 z_1)^2
+ 0.552 \exp(0.510 z_2)
+ 0.617 \sin(1.208 z_3 - 2.997)
- 0.958.
\label{eq:nl_trans}
\end{align}
\end{subequations}

All fitted representations recover the correct qualitative structure of the data-generating mechanism: quadratic in $z_1$, exponential in $z_2$, and sinusoidal in $z_3$, up to affine rescaling and additive constants. Differences in numerical coefficients are expected, as the nonlinear AFT regression function is identifiable only up to equivalent reparameterizations and is estimated under noise and right-censoring.

Importantly, the KAN-AFT framework achieves accurate structural recovery of nonlinear covariate effects rather than exact coefficient matching, demonstrating its ability to adaptively capture complex functional relationships while retaining interpretability.
\begin{table}[h!tbp]
\centering
\caption{Comparison of AFT estimators on nonlinear synthetic data}
\label{tab:model_comparison2}
\begin{tabular}{@{}l cccc@{}}
\toprule
Model & \multicolumn{2}{c}{C-index} & \multicolumn{2}{c}{MSE}\\
\cmidrule(r){2-3}\cmidrule(l){4-5}
& Train & Test & Train & Test\\
\midrule
AFT-Weibull & 0.7387 & 0.7251 & 6.9738 & 7.2768\\
AFT-Log-normal & 0.7415 & 0.7293 & 6.4513 & 6.7199\\
AFT-Log-logistic & 0.7432 & 0.7328 & 6.4807 & 6.7034\\
\midrule
DeepAFT-BuckleyJames &  0.7477  & 0.7228 & 6.2107 & 6.4105\\
DeepAFT-IPCW & 0.7892  & 0.7797  & 4.0080& 4.3852\\
DeepAFT-Transform & 0.7691  & 0.7560  & 5.8764 & 6.1042\\
\midrule
KAN-AFT-BuckleyJames & \textbf{0.8505} & \textbf{0.8409} & \textbf{0.2781} & 0.2948\\
KAN-AFT-IPCW & 0.8317 & 0.8281 & 0.4354 & 0.4087\\
KAN-AFT-Transform & 0.8430 & 0.8373 & 0.2865& \textbf{0.2751}\\
\bottomrule
\end{tabular}
\end{table}

Table~\ref{tab:model_comparison2} reports predictive performance. Parametric AFT models perform poorly under this nonlinear specification due to model misspecification, while DeepAFT exhibits moderate improvement. In contrast, KAN-AFT attains substantially higher C-index values and lower MSE across all censoring strategies.
%%%%%%%%%%%%%%%%%%%%%%%%%%%%%%%%%%%%%%%%%%%%%%%
\subsection{Veteran Lung Cancer Data}

The Veteran lung cancer dataset \cite{kalbfleisch2002statistical} arises from a randomized clinical trial comparing two treatment regimens for advanced lung cancer and contains survival information for $n=137$ patients, with approximately 7\% right-censored observations. The outcome of interest is survival time (in days) from initiation of treatment. We consider three covariates: age ($z_1$), Karnofsky performance score ($z_2$), and months from diagnosis to randomization ($z_3$).

Schoenfeld residual diagnostics are reported in Table~\ref{tab:ph_check_veteran}. Small p-values suggest a departure from the proportional hazards assumption. The proportional hazards assumption is violated for the Karnofsky performance score and at the global level, motivating the use of AFT–based models for this dataset.

\begin{table}[!htbp]
\centering
\caption{Schoenfeld residual test for proportional hazards assumption in the Veteran lung cancer dataset.}
\label{tab:ph_check_veteran}
\begin{tabular}{lcc}
\toprule
Covariate & $\chi^{2}$ statistic & p-value \\
\midrule
Age ($z_1$) & 2.135 & 0.1439 \\
Karnofsky score ($z_2$) & 13.096 & 0.0003 \\
Diagnosis time ($z_3$) & 0.008 & 0.9304 \\
\midrule
Global test & 17.805 & 0.0005 \\
\bottomrule
\end{tabular}
\end{table}

Equations~\eqref{eq:veteran_bj}--\eqref{eq:veteran_trans} present the symbolic approximations obtained under the Buckley--James, IPCW, and transformation-based KAN-AFT fits, respectively.
\vspace{-.3cm}
\begin{subequations}
\label{eq:veteran_symbolic_models}
\begin{align}
\log(\hat{T}) &=
0.002(-5.028\, z_1 - 6.44)^2
+ 0.840\, z_2
+ 1.37\times 10^{-5}(-6.808\, z_3 - 9.960)^2
- 0.097,
\label{eq:veteran_bj}\\[4pt]
\log(\hat{T}) &=
0.002(-6.259\, z_1 - 8.018)^2
+ 0.740\, z_2
+ 1.15\times 10^{-4}(-5.265\, z_3 - 7.69)^2
- 0.290,
\label{eq:veteran_ipcw}\\[4pt]
\log(\hat{T}) &=
0.168\, z_1
+ 0.730\, z_2
+ 2.14\times 10^{-5}(-6.48\, z_3 - 9.48)^2
- 0.017.
\label{eq:veteran_trans}
\end{align}
\end{subequations}

Table~\ref{tab:model_comparison5} shows that KAN-AFT with Buckley--James imputation attains the highest test-set C-index, indicating superior ranking performance under right-censoring. While parametric AFT models perform competitively, KAN-AFT provides improved robustness by relaxing linearity and distributional assumptions and by yielding transparent covariate-specific effect shapes.

\begin{table}[!htbp]
\centering
\caption{Predictive performance (C-index) of AFT estimators on the Veteran lung cancer dataset.}
\label{tab:model_comparison5}
\begin{tabular}{lcc}
\toprule
Model & Training C-index & Test C-index \\
\midrule
AFT--Weibull        & 0.7064 & 0.7285 \\
AFT--Log-normal    & 0.7113 & 0.7064 \\
AFT--Log-logistic  & 0.7111 & 0.7091 \\
\midrule
DeepAFT--Buckley--James & 0.7124 & 0.6884 \\
DeepAFT--IPCW          & 0.7179 & 0.6911 \\
DeepAFT--Transform     & --     & --     \\
\midrule
KAN-AFT--Buckley--James & 0.7124 & \textbf{0.7313} \\
KAN-AFT--IPCW           & 0.7091 & 0.7091 \\
KAN-AFT--Transform      & 0.7117 & 0.7147 \\
\bottomrule
\end{tabular}
\end{table}

Inspection of the Buckley--James symbolic representation in Equation~\eqref{eq:veteran_bj} reveals a simple additive structure in which the dominant contribution to $\log(\hat T)$ arises from the Karnofsky performance score ($z_2$), which enters linearly with a large positive coefficient. This indicates a strong monotone effect whereby improved functional status is associated with substantially longer survival, identifying $z_2$ as the primary prognostic factor in the model. Age ($z_1$) enters through a quadratic transformation with a comparatively small coefficient, indicating a weak but nonlinear association in which survival varies smoothly across age levels rather than following a strictly monotone trend. Months from diagnosis to randomization ($z_3$) appears only through a very low-magnitude quadratic term, suggesting a negligible independent contribution once functional status and age are accounted for. This indicates that survival in this dataset is driven predominantly by functional performance, with age acting as a secondary nonlinear modifier and diagnosis time exerting minimal influence. These findings are consistent with the strong violation of the proportional hazards assumption for the Karnofsky score and with established analyses of the Veteran lung cancer data \cite{kalbfleisch2002statistical}.

\begin{figure}[!htbp]
    \centering
    \includegraphics[width=1\linewidth]{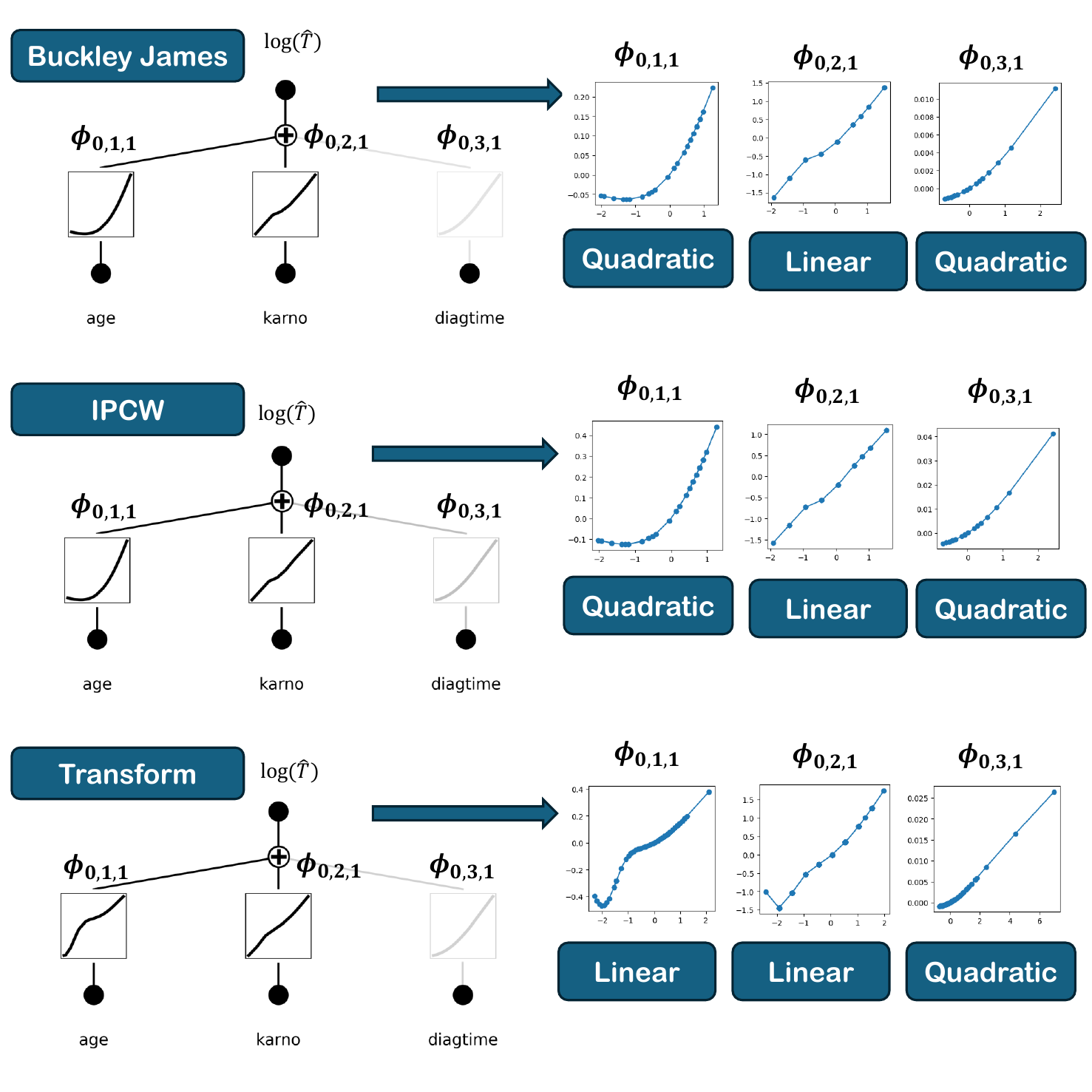}
    \caption{Structured KAN-AFT representation for the Veteran lung cancer dataset.}
    \label{fig:veteran_network}
\end{figure}

Figure~\ref{fig:veteran_network} illustrates the KAN-AFT representation for this dataset. Examination of the pruned KAN-AFT model without hidden interactions reveals a structurally simple formulation with transparent and clinically interpretable covariate effects.
%%%%%%%%%%%%%%%%%%%%%%%%%%%%%%%%%%%%%%%%%%%%%%
\subsection{German Breast Cancer Study Group (GBSG) Data}
The German Breast Cancer Study Group (GBSG) dataset arises from a multicenter clinical trial of patients with node-positive breast cancer \cite{royston2013external}. The analysis retains 686 patients with complete covariate information, among whom approximately 56\% are right-censored. The outcome of interest is recurrence-free survival time, defined as the time to first recurrence, death, or last follow-up. We consider five prognostic covariates: age ($z_1$), tumor size ($z_2$), number of positive lymph nodes ($z_3$), progesterone receptor level ($z_4$), and estrogen receptor level ($z_5$).

Schoenfeld residual diagnostics for the GBSG dataset are reported in Table~\ref{tab:ph_check_gbsg}. Significant violations of the proportional hazards assumption are observed for age, progesterone receptor, estrogen receptor, and at the global level, motivating the use of AFT-based models for this dataset.

\begin{table}[!htbp]
\centering
\caption{Schoenfeld residual test for proportional hazards assumption in the GBSG dataset.}
\label{tab:ph_check_gbsg}
\begin{tabular}{lcc}
\toprule
Covariate & $\chi^{2}$ statistic & p-value \\
\midrule
Age ($z_1$) & 11.582 & 0.00067 \\
Tumor size ($z_2$) & 0.120 & 0.72949 \\
Nodes ($z_3$) & 0.897 & 0.34347 \\
Progesterone receptor ($z_4$) & 4.954 & 0.02603 \\
Estrogen receptor ($z_5$) & 6.599 & 0.01020 \\
\midrule
Global test & 18.448 & 0.00243 \\
\bottomrule
\end{tabular}
\end{table}

Equations~\eqref{eq:gbsg_bj}--\eqref{eq:gbsg_transform} present the symbolic approximations corresponding to the Buckley--James, IPCW, and transformation-based KAN-AFT fits, respectively.
\vspace{-.3cm}
\begin{subequations}
\label{eq:gbsg_symbolic_models}
\begin{align}
\log(\hat{T}) &=
0.0002(-5.6756\, z_1 - 7.5027)^2
- 0.0002(-5.7468\, z_2 - 7.5802)^2
\nonumber\\
&\quad
- 0.0016(-4.6948\, z_3 - 6.5282)^2
+ 0.1179\, z_4
+ 0.0548\, z_5
+ 0.0917,
\label{eq:gbsg_bj}\\[4pt]
\log(\hat{T}) &=
- 0.0003(-7.1999\, z_3 - 9.9989)^2
- 0.1479 \exp\!\left(-(-1.4629\, z_4 - 0.5791)^2\right)
\nonumber\\
&\quad
+ 0.0545\, z_5
- 0.0354,
\label{eq:gbsg_ipcw}\\[4pt]
\log(\hat{T}) &=
- 0.0001(-5.4763\, z_2 - 7.2239)^2
- 0.0004(-7.1960\, z_3 - 9.9933)^2
\nonumber\\
&\quad
+ 0.0251\, z_4
+ 0.0466\, z_5
+ 0.0549.
\label{eq:gbsg_transform}
\end{align}
\end{subequations}

Table~\ref{tab:model_comparison4} reports the predictive performance of parametric AFT, DeepAFT, and KAN-AFT models on the GBSG dataset. Parametric specifications attain the highest test-set C-index values, indicating that a relatively simple parametric AFT structure is adequate for this dataset. Nevertheless, the proposed KAN-AFT model achieves test-set C-index values that are close to those of the best-performing parametric AFT models and substantially higher than those obtained by DeepAFT under all censoring strategies.
\begin{table}[!htbp]
\centering
\caption{Predictive performance (C-index) of AFT estimators on the GBSG dataset.}
\label{tab:model_comparison4}
\begin{tabular}{lcc}
\toprule
Model & Training C-index & Test C-index \\
\midrule
AFT--Weibull        & 0.6756 & 0.7088 \\
AFT--Log-normal    & 0.6759 & 0.6999 \\
AFT--Log-logistic  & 0.6762 & 0.7004 \\
\midrule
DeepAFT--Buckley--James & 0.5427 & 0.5200 \\
DeepAFT--IPCW          & 0.5446 & 0.5375 \\
DeepAFT--Transform     & 0.5532 & 0.5378 \\
\midrule
KAN-AFT--Buckley--James & 0.6598 & 0.6812 \\
KAN-AFT--IPCW           & 0.6534 & 0.6648 \\
KAN-AFT--Transform      & 0.6587 & 0.6934 \\
\bottomrule
\end{tabular}
\end{table}

Inspection of the Buckley--James symbolic representation in Equation~\eqref{eq:gbsg_bj} reveals a parsimonious additive structure dominated by nodal involvement and hormone receptor status. The number of positive lymph nodes ($z_3$) enters through a negative quadratic term with the largest magnitude, identifying nodal burden as the primary adverse prognostic factor for recurrence-free survival. Progesterone receptor ($z_4$) and estrogen receptor ($z_5$) appear as positive linear terms, indicating improved prognosis with increasing hormone receptor expression. Age ($z_1$) and tumor size ($z_2$) enter through low-amplitude quadratic components in the Buckley--James fit and are pruned in the IPCW model, suggesting comparatively weak independent effects once nodal status and receptor expression are accounted for. The model highlights nodal burden as the dominant driver of recurrence risk, with hormone receptor levels providing strong protective effects, consistent with established clinical understanding of breast cancer prognosis.

\begin{figure}[!htbp]
\centering
\includegraphics[width=0.8\linewidth]{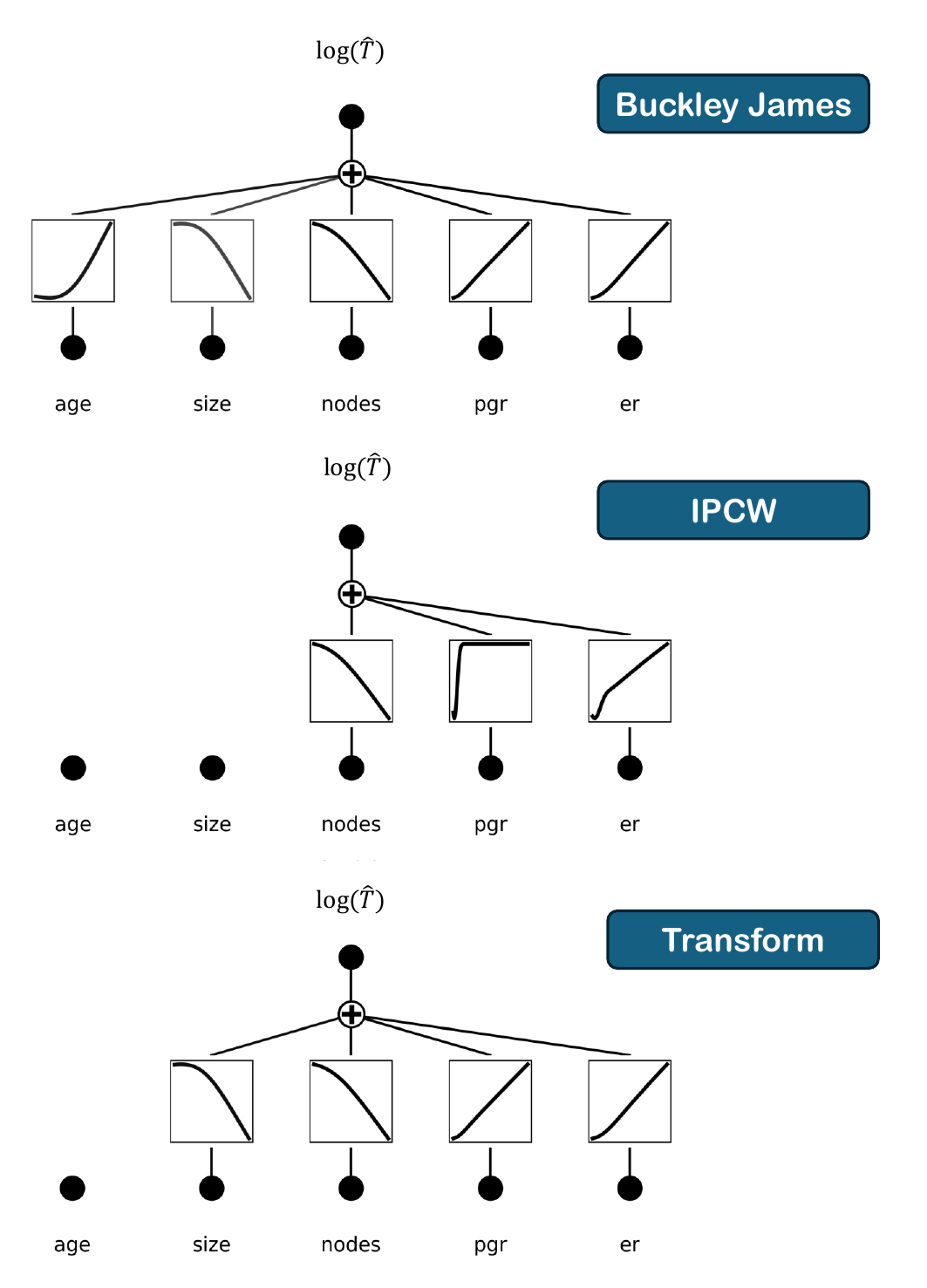}
\caption{Structured KAN-AFT representation for the GBSG dataset.}
\label{fig:gbsg_network}
\end{figure}

Figure~\ref{fig:gbsg_network} illustrates the KAN-AFT representation for this dataset. This example further illustrates that when covariate effects are largely smooth and simple, the proposed KAN-AFT framework naturally adapts by selecting a parsimonious structure while maintaining competitive predictive performance.

%%%%%%%%%%%%%%%%%%%%%%%%%%%%%%%%%%%%%%%%%%%%%
\subsection{Primary Biliary Cholangitis (PBC) Data}
The Mayo Clinic Primary Biliary Cholangitis (PBC) dataset \cite{therneau2000cox} contains clinical information on 418 patients diagnosed with primary biliary cholangitis, an autoimmune liver disease, with approximately 53\% right-censored observations after removal of incomplete records. The outcome of interest is time to death or liver transplantation. We consider ten prognostic covariates: age ($z_1$), serum bilirubin ($z_2$), serum cholesterol ($z_3$), albumin ($z_4$), urine copper ($z_5$), alkaline phosphatase ($z_6$), aspartate aminotransferase (AST; $z_7$), triglycerides ($z_8$), platelet count ($z_9$), and prothrombin time ($z_{10}$).

Schoenfeld residual diagnostics are reported in Table~\ref{tab:ph_check_pbc}. Several covariates, including bilirubin, cholesterol, triglycerides, platelet count, and prothrombin time, exhibit significant departures from the proportional hazards assumption, and the global test is also significant. These results motivate the use of AFT–based models for this dataset.

\begin{table}[!htbp]
\centering
\caption{Schoenfeld residual test for proportional hazards assumption in the PBC dataset.}
\label{tab:ph_check_pbc}
\begin{tabular}{lcc}
\toprule
Covariate & $\chi^{2}$ statistic & p-value \\
\midrule
Age ($z_1$) & 2.52 & 0.1122 \\
Bilirubin ($z_2$) & 10.50 & 0.0012 \\
Cholesterol ($z_3$) & 9.69 & 0.0019 \\
Albumin ($z_4$) & 1.57 & 0.2095 \\
Copper ($z_5$) & 0.42 & 0.5171 \\
Alkaline phosphatase ($z_6$) & 1.53 & 0.2160 \\
AST ($z_7$) & 1.35 & 0.2450 \\
Triglycerides ($z_8$) & 5.98 & 0.0145 \\
Platelet count ($z_9$) & 3.94 & 0.0472 \\
Prothrombin time ($z_{10}$) & 4.51 & 0.0338 \\
\midrule
Global test & 26.90 & 0.0027 \\
\bottomrule
\end{tabular}
\end{table}

Equations~\eqref{eq:pbc_bj}--\eqref{eq:pbc_transform} present the symbolic approximations obtained respectively under the Buckley--James, IPCW, and transformation-based KAN-AFT fits.
\vspace{-.3cm}
\begin{subequations}
\label{eq:pbc_symbolic_models}
\begin{align}
\log(\hat{T}) &=
- 1.331 \sin(0.303\, z_1 - 7.477)
- 1.256 \sin(0.388\, z_2 - 7.402)
\nonumber\\
&\quad
- 0.101 \sin(0.357\, z_3 - 1.138)
- 0.350 \cos(0.357\, z_3 + 0.433)
\nonumber\\
&\quad
+ 0.002(-4.983\, z_4 - 7.983)^2
\nonumber\\
&\quad
+ 0.002\!\left(-5.655 \sin(0.369\, z_5 + 2.012) - 2.883\right)^2
- 0.671 \cos(0.368\, z_5 - 8.983)
\nonumber\\
&\quad
+ 0.105 \sin(0.325\, z_8 - 1.150)
- 0.472 \cos(0.377\, z_9 - 5.828)
\nonumber\\
&\quad
+ 0.603 \sin(0.360\, z_{10} + 1.995)
- 2.730,
\label{eq:pbc_bj}\\[4pt]
\log(\hat{T}) &=
0.008\!\left(
6.355 \cos(0.466\, z_2 + 3.608)
- 3.121 \sin(0.461\, z_{10} - 7.377)
- 2.646
\right)^2
\nonumber\\
&\quad
- 1.102 \sin(0.385\, z_1 - 7.401)
- 0.469 \sin(0.466\, z_2 - 7.388)
\nonumber\\
&\quad
- 0.677 \sin(0.368\, z_3 + 2.021)
+ 0.462 \sin(0.370\, z_5 + 2.017)
\nonumber\\
&\quad
- 0.205 \cos(0.405\, z_6 + 3.583)
+ 0.200 \sin(0.475\, z_8 - 7.372)
\nonumber\\
&\quad
+ 0.325 \sin(0.358\, z_9 + 5.157)
- 0.640 \sin(0.461\, z_{10} - 7.378)
- 1.841,
\label{eq:pbc_ipcw}\\[4pt]
\log(\hat{T}) &=
\phi_{1,1,1}(x)
- 0.722 \sin(0.387\, z_1 - 7.398)
- 0.460 \sin(0.466\, z_2 - 7.387)
\nonumber\\
&\quad
- 0.255 \sin(0.369\, z_3 + 2.021)
+ 0.543 \sin(0.370\, z_5 + 2.017)
\nonumber\\
&\quad
+ 0.186 \sin(0.475\, z_8 - 7.372)
+ 0.786 \sin(0.353\, z_9 + 5.152)
\nonumber\\
&\quad
- 0.551 \sin(0.461\, z_{10} - 7.378)
- 0.905,
\nonumber\\
&\quad
\text{where, } \phi_{1,1,1}(x) =  \left(-x - 2.6495\right)
\sin\!\left(\frac{x}{-4.2497}\right).
\label{eq:pbc_transform}
%\phi_{1,1,1}(x) &=
%\left(-x - 2.6495\right)
%\sin\!\left(\frac{x}{-4.2497}\right).
%\label{eq:pbc_hidden}
\end{align}
\end{subequations}

Here $x$ denotes the aggregated hidden-layer input, $x=\sum_{j}\phi_{0,j,1}(z_j)$, so that $\phi_{1,1,1}(\cdot)$ represents a nonlinear calibration applied to the combined covariate effect rather than to any single covariate.

Table~\ref{tab:model_comparison3} shows that KAN-AFT with Buckley--James imputation attains the highest test-set C-index. Across censoring strategies, KAN-AFT consistently outperforms DeepAFT and achieves performance comparable to, or exceeding, parametric AFT models, while additionally providing transparent nonlinear covariate-effect representations.

\begin{table}[!htbp]
\centering
\caption{Predictive performance (C-index) of AFT estimators on the PBC dataset.}
\label{tab:model_comparison3}
\begin{tabular}{lcc}
\toprule
Model & Training C-index & Test C-index \\
\midrule
AFT--Weibull        & 0.8058 & 0.7958 \\
AFT--Log-normal    & 0.8046 & 0.7993 \\
AFT--Log-logistic  & 0.8045 & 0.7958 \\
\midrule
DeepAFT--Buckley--James & 0.7504 & 0.7729 \\
DeepAFT--IPCW          & 0.6595 & 0.6474 \\
DeepAFT--Transform     & 0.7053 & 0.6955 \\
\midrule
KAN-AFT--Buckley--James & 0.8040 & \textbf{0.8142} \\
KAN-AFT--IPCW           & 0.7751 & 0.7901 \\
KAN-AFT--Transform      & 0.8008 & 0.7833 \\
\bottomrule
\end{tabular}
\end{table}

Inspection of the Buckley--James symbolic representation in Equation \eqref{eq:pbc_bj} shows that the dominant contributions to $\log(\hat T)$ arise from covariates $z_1$ (age) and $z_2$ (serum bilirubin), each entering through single smooth trigonometric components with comparatively large magnitudes. These large-amplitude sine terms indicate that both age and bilirubin exert strong nonlinear effects on survival, with increasing values shifting the latent index toward regions associated with lower log-survival time. Serum cholesterol ($z_3$) appears through a combination of sine and cosine components with moderate coefficients, implying a smooth but weaker nonlinear influence relative to $z_1$ and $z_2$. Albumin ($z_4$) enters via a quadratic transformation with a small positive coefficient, reflecting a smooth protective effect with diminishing marginal gains at higher albumin levels. Urine copper ($z_5$) contributes through both a quadratic modulation and a cosine component, indicating a nonlinear adverse effect that saturates at extreme values. Triglycerides ($z_8$) enter through a low-amplitude sine term, suggesting a secondary role in determining survival. Platelet count ($z_9$) is associated with a moderate-magnitude cosine component, consistent with the adverse prognostic impact of thrombocytopenia. Prothrombin time ($z_{10}$) appears through a sine term, indicating a clinically meaningful nonlinear effect consistent with impaired coagulation and advanced liver dysfunction. Notably, alkaline phosphatase ($z_6$) and AST ($z_7$) are pruned from the Buckley--James fit, implying negligible independent contribution once the remaining biochemical markers are accounted for.

The model structure suggests that survival in PBC is driven primarily by age ($z_1$) and serum bilirubin ($z_2$), with additional modulation from albumin ($z_4$), urine copper ($z_5$), platelet count ($z_9$), and prothrombin time ($z_{10}$), while alkaline phosphatase ($z_6$) and AST ($z_7$) contribute little independent information. This covariate importance ordering is broadly consistent with the classical Mayo PBC risk model and related analyses \cite{therneau2000cox}.

\begin{figure}[!htbp]
\centering
\includegraphics[width=1\linewidth]{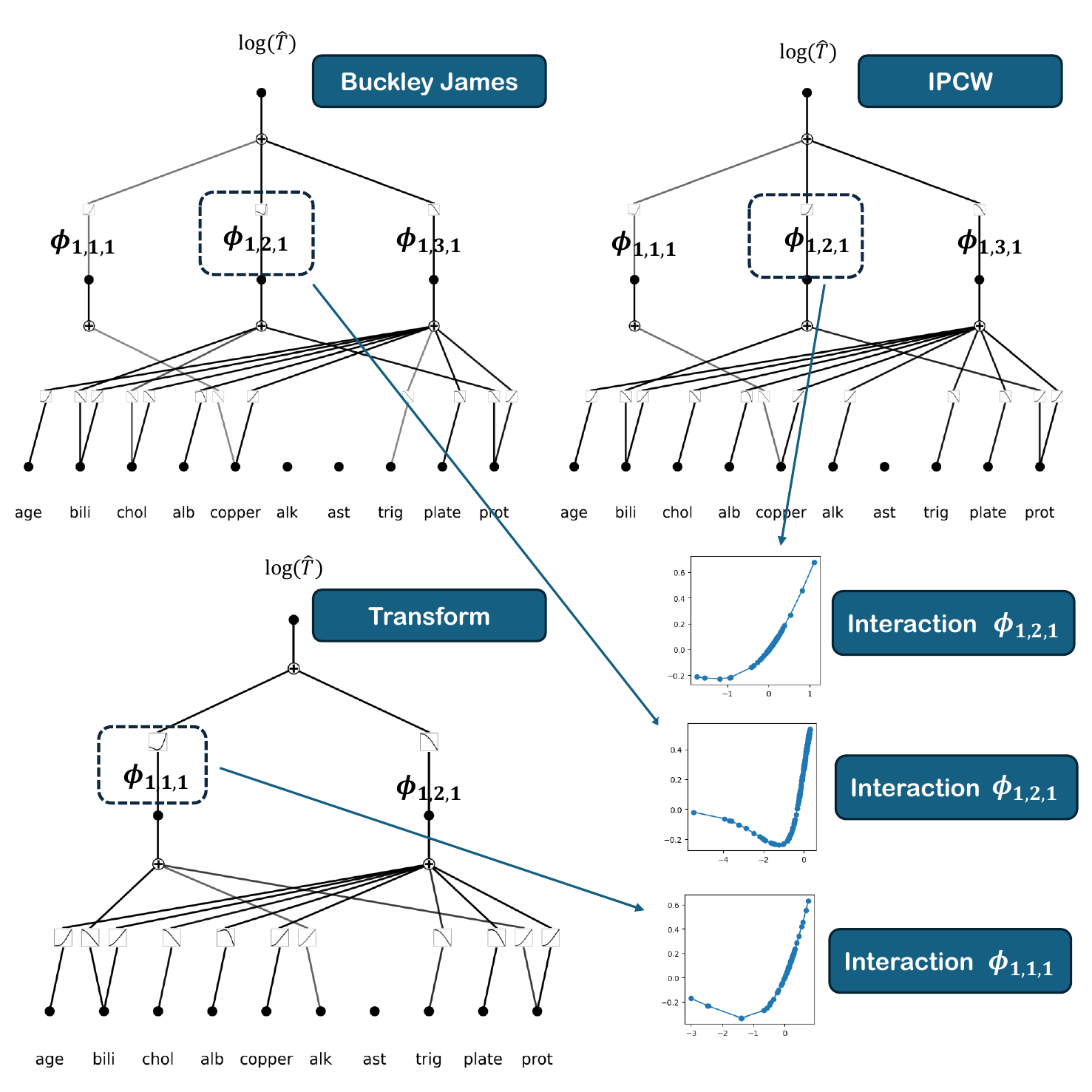}
\caption{Structured KAN-AFT representation for the PBC dataset.}
\label{fig:pbc_network}
\end{figure}

Figure~\ref{fig:pbc_network} illustrates the KAN-AFT representation for this dataset. Although the symbolic approximations involve trigonometric basis functions, the corresponding learned univariate components exhibit smooth, largely monotone or threshold-type shapes over the observed covariate ranges. Thus, these terms should be interpreted as flexible smooth approximations rather than as evidence of true periodic biological mechanisms.

%%%%%%%%%%%%%%%%%%%%%%%%%%%%%%%%%%%%%%%%%%%%%%%%%%%%%%%%%%
\subsection{Heart Failure Clinical Records Data}

The Heart Failure Clinical Records dataset \cite{chicco2020machine} contains longitudinal information on 299 patients diagnosed with heart failure, with approximately 68\% of observations subject to right-censoring. The outcome of interest is time to death due to heart failure. We consider six clinically relevant covariates: age ($z_1$), creatinine phosphokinase ($z_2$), ejection fraction ($z_3$), platelet count ($z_4$), serum creatinine ($z_5$), and serum sodium ($z_6$).

\begin{figure}[!htbp]
    \centering
    \includegraphics[width=1\linewidth]{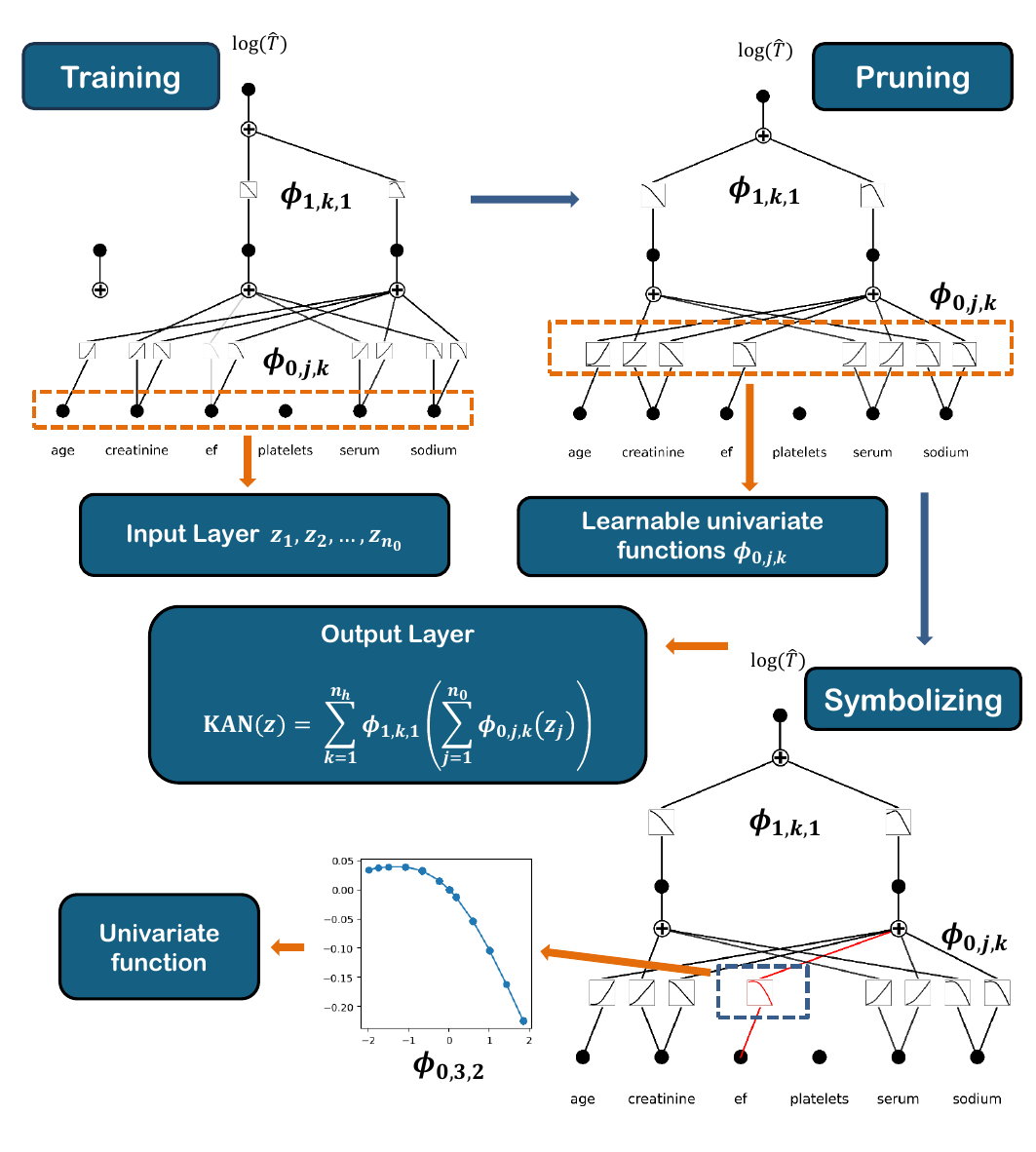}
    \caption{Structured KAN-AFT modeling pipeline applied to the Heart Failure dataset.}
    \label{fig:heart_pipeline}
\end{figure}

Figure~\ref{fig:heart_pipeline} illustrates the structured estimation pipeline of KAN-AFT, comprising training of univariate edge functions, sparsity-inducing pruning, and symbolic approximation. Equations~\eqref{eq:heart_bj}--\eqref{eq:heart_trans} present the symbolic approximations obtained under the Buckley--James, IPCW, and transformation-based KAN-AFT fits, respectively.

\vspace{-1cm}
\begin{subequations}
\label{eq:heart_symbolic_models}
\begin{align}
\log(\hat{T}) &=
-0.0038
\Bigl(
0.0040(-7.0626\, z_6 - 9.4790)^2
- 6.0886 \sin(0.4505\, z_2 - 7.4185)
\nonumber\\
&\quad
+ 1.4180 \cos(0.5260\, z_5 + 6.7696)
- 14.4968
\Bigr)^2
\nonumber\\
&\quad
+ 1.2650 \cos\Bigl(
1.4002 \cos(0.3902\, z_1 - 5.8161)
+ 0.7494 \sin(0.4496\, z_2 - 7.4182)
\nonumber\\
&\quad\quad
+ 0.0005(-6.8905\, z_3 - 8.6681)^2
+ 0.1331 \sin(0.5386\, z_5 - 4.2229)
\nonumber\\
&\quad\quad
+ 0.0008(-7.1317\, z_6 - 9.5680)^2
+ 4.7440
\Bigr)
- 0.4753,
\label{eq:heart_bj}\\[4pt]
\log(\hat{T}) &=
-0.0043
\Bigl(
- 3.3001 \sin(0.3907\, z_1 + 5.1802)
+ 5.0622 \sin(0.4496\, z_2 - 7.4178)
- 4.3440
\Bigr)^2
\nonumber\\
&\quad
+ 0.5486 \sin\Bigl(
1.1127 \cos(0.3924\, z_1 - 5.8138)
+ 1.4473 \sin(0.4498\, z_2 - 7.4182)
\nonumber\\
&\quad\quad
- 0.0010(-5.3534\, z_3 - 6.7313)^2
+ 0.3155 \cos(0.4178\, z_4 - 5.7958)
\nonumber\\
&\quad\quad
- 1.6492 \cos(0.5110\, z_5 - 8.9444)
+ 0.0005(-7.0926\, z_6 - 9.5162)^2
- 1.6123
\Bigr)
\nonumber\\
&\quad
+ 1.3977 \cos\Bigl(
0.3398 \sin(0.3915\, z_1 + 5.1810)
+ 0.5987 \sin(0.4496\, z_2 - 7.4181)
\nonumber\\
&\quad\quad
- 0.0005(-6.8965\, z_6 - 9.2546)^2
+ 7.6565
\Bigr)
- 1.2173,
\label{eq:heart_ipcw}\\[4pt]
\log(\hat{T}) &=
0.2437
- 0.0038
\Bigl(
1.2662 \sin(0.3996\, z_1 + 5.1880)
- 6.0283 \sin(0.4499\, z_2 - 7.4183)
\nonumber\\
&\quad
- 1.4292 \cos(0.4020\, z_4 - 5.8093)
+ 0.6694 \sin(0.5572\, z_5 - 4.2166)
- 11.5618
\Bigr)^2.
\label{eq:heart_trans}
\end{align}
\end{subequations}

\begin{table}[!htbp]
\centering
\caption{Predictive performance (C-index) of AFT estimators on the Heart Failure Clinical Records dataset.}
\label{tab:model_comparison6}
\begin{tabular}{lcc}
\toprule
Model & Training C-index & Test C-index \\
\midrule
AFT--Weibull        & 0.7149 & 0.7593 \\
AFT--Log-normal    & 0.7145 & 0.7593 \\
AFT--Log-logistic  & \textbf{0.7155} & 0.7581 \\
\midrule
DeepAFT--Buckley--James & 0.6800 & 0.7182 \\
DeepAFT--IPCW          & 0.5339 & 0.5262 \\
DeepAFT--Transform     & 0.5403 & 0.5461 \\
\midrule
KAN-AFT--Buckley--James & 0.6703 & \textbf{0.7618} \\
KAN-AFT--IPCW           & 0.6825 & 0.7170 \\
KAN-AFT--Transform      & 0.7011 & 0.7382 \\
\bottomrule
\end{tabular}
\end{table}

Table~\ref{tab:model_comparison6} summarizes predictive performance. Across all censoring strategies, KAN-AFT consistently outperforms DeepAFT on the test set and achieves performance comparable to, or exceeding, parametric AFT models. In particular, KAN-AFT with Buckley--James imputation attains the highest test-set C-index.

For ease of interpretation, the symbolic representation in Equation~\eqref{eq:heart_bj} can be viewed through a simplified interaction structure obtained by neglecting the weaker quadratic block while retaining the dominant cosine block and constant term. Under this approximation,
\[
\log(\hat T) \approx 1.2650 \cos(B) - 0.4753,
\]
where $B$ denotes a nonlinear additive index of the covariates. Covariates therefore influence survival primarily by shifting the latent index $B$. Within the observed clinical range, $B$ lies in a locally monotone decreasing region of the cosine transformation, such that increases in physiological risk markers that raise $B$ correspond to shorter predicted survival times, whereas decreases in $B$ correspond to longer survival expectations.

Serum creatinine ($z_5$) appears as the dominant adverse prognostic factor, entering through high-magnitude trigonometric components that ensure marginal increases in levels lead to substantial positive shifts in $B$ and sharp reductions in $\hat{T}$. Ejection fraction ($z_3$) and serum sodium ($z_6$) contribute through smooth quadratic terms, capturing protective effects and the risk associated with hyponatremia, respectively. The quadratic nature of the ejection fraction term specifically reflects a diminishing marginal benefit, where survival gains plateau once cardiac efficiency reaches a certain threshold. Creatinine phosphokinase ($z_2$) is represented by bounded sinusoidal components, reflecting a secondary association where the impact of enzyme spikes is capped by the periodic nature of the function, preventing outliers from dominating the global prediction. Age ($z_1$) enters via a smooth nonlinear cosine transformation that acts as a progressive driver of the latent index, steadily moving the phase toward the cosine trough with advancing age. Platelet count ($z_4$) is fully pruned from the model, indicating a lack of independent prognostic value.

These findings closely mirror prior analyses of the same dataset, which consistently identify serum creatinine and ejection fraction as the most influential predictors of survival, with age and serum sodium also exhibiting clear associations with mortality, while CPK shows a weaker secondary effect and platelet count provides limited independent prognostic information \cite{chicco2020machine,srujana2024machine}.

The simulation experiments demonstrate that the proposed KAN-AFT framework adapts its structural complexity to the underlying data-generating mechanism. In the linear setting, the model collapses to a parsimonious additive form without introducing spurious nonlinearities. Under nonlinear specifications, it accurately recovers the qualitative functional structure of covariate effects. Across multiple real-world survival datasets with varying degrees of censoring and structural complexity, KAN-AFT attains competitive or superior predictive performance while yielding transparent symbolic representations of covariate effects. Collectively, these results establish that the proposed approach achieves a favorable balance between flexibility, interpretability, and predictive accuracy in censored survival settings.
%%%%%%%%%%%%%%%%%%%%%%%%%%%%%%%%%%%%%%%%%%%%%%%%
\section{Conclusion}
\label{Section:Conclusion}

This work develops a structured nonparametric formulation of accelerated failure time regression that extends the classical AFT paradigm beyond linear predictors while preserving its fundamental time-scale interpretation. By expressing the AFT regression function through compositions of univariate smooth components, the proposed framework provides a practical compromise between model flexibility and interpretability, two features that are often in tension in modern lifetime data analysis .

From a modeling perspective, the framework unifies classical AFT regression and flexible nonlinear regression within a single formulation. The linear AFT model arises as a special case under simple structural restrictions on the component functions, clarifying the connection between the proposed approach and standard methodology. At the same time, the structured representation enables nonlinear covariate effects and limited interaction patterns to be captured in a transparent manner, without requiring ad hoc basis specification or manual construction of nonlinear terms.

The methodology is designed for direct use with right-censored data and accommodates several established censoring-adjusted estimation strategies. This modularity allows analysts to select estimation procedures according to data characteristics and computational considerations, while retaining a common regression representation. The resulting fitted models admit covariate-specific effect functions that can be visualized or expressed symbolically, facilitating substantive interpretation in applied studies.

The framework is broadly applicable to time-to-event problems in clinical research, epidemiology, reliability engineering, and related areas where interest centers on how covariates modify event timing rather than instantaneous risk. In such settings, the ability to obtain smooth, interpretable covariate effects while avoiding restrictive linear assumptions can provide meaningful scientific insight in addition to accurate prediction.

The preceding analysis naturally opens up several questions that warrant further investigation. The current focus on a single event type and independent censoring, for instance, invites methodological extensions to more intricate observational settings. Adapting the estimation procedure to accommodate competing risks or multi-state processes would not only broaden the framework's applicability but also test the stability of the discovered symbolic forms under more complex time-to-event settings.

In summary, this work contributes a flexible and interpretable extension of accelerated failure time modeling that remains closely connected to classical lifetime data analysis  while addressing the demands of contemporary applied problems.

%%%%%%%%%%%%%%%%%%%%%%%%%%%%%%%%%%%%%%%%%%%%%%%%%%%%%%%%%%%%%%%%%%%%%%%%%%%%%%%%%
\section*{Data Availability Statement}

The authors confirm that the data supporting the findings of this study are available within the article. All results can be reproduced from the equations and methodological details provided in the manuscript. Where external data sources are used, the appropriate references have been cited and the data are publicly available.
%%%%%%%%%%%%%%%%%%%%%%%%%%%%%%%%%%%%%%%%%%%%%%%%%%%%%%%%%%%%%%%%%%%%%%%%%%%%%%%%%
\appendix
\section{Censoring Mechanisms}
\label{Append_Censoring}

The accelerated failure time (AFT) model aims to estimate the conditional mean of the
log-failure time. Let $T_i$ denote the latent event time and $C_i$ the censoring time
for subject $i$. The observed data are
\(
\tilde T_i = \min(T_i, C_i),
\quad
\delta_i = \mathbb I(T_i \le C_i),
\quad
\mathbf z_i \in \mathbb R^{n_{\mathrm{cov}}}.
\)
 We assume that conditional on the covariates, the failure time $T_i$ and the censoring time $C_i$ are independent, that is,
$C_i \perp T_i \mid \mathbf z_i$.

Throughout this appendix, the nonlinear AFT model is specified by
$$\log T_i = \eta(\mathbf z_i) + \varepsilon_i,
\quad
\mathbb E(\varepsilon_i \mid \mathbf z_i)=0,$$
where $\eta(\cdot)$ is approximated by  univariate functions,
$\eta(\mathbf z) \approx \mathrm{KAN}(\mathbf z)$.

%%%%%%%%%%%%%%%%%%%%%%%%%%%%%%%%%%%%%%%%%%%%%%%%%%%%%%%%%%%%%%
\subsection{KAN-AFT--Buckley--James (Iterative Imputation)}

The Buckley--James approach \cite{BuckleyJames1979} extends least-squares regression to right-censored data through iterative imputation of conditional residuals. Let \(Y_i=\log(\tilde T_i)\) denote the working response. The Buckley--James KAN--AFT estimator is obtained via the following iterative procedure:

\begin{enumerate}
\item {Initialization:}
Set $\hat\eta^{(0)}(\mathbf z_i)=0$ for all $i$.

\item {Residual computation (iteration $k$):}
\(
r_i^{(k-1)} =
\tilde T_i \exp\{-\hat\eta^{(k-1)}(\mathbf z_i)\}.
\)

\item {Estimate residual survival:}
Apply the Kaplan--Meier estimator to $\{r_i^{(k-1)}\}$.

\item {Imputation:}
\[
r_i^{*(k-1)}=
\begin{cases}
r_i^{(k-1)}, & \delta_i=1,\\[4pt]
\mathbb E\!\left(r \mid r>r_i^{(k-1)}\right), & \delta_i=0 .
\end{cases}
\]

\item {Updated pseudo-response:}
\(
Y_i^{*(k)}=
\hat\eta^{(k-1)}(\mathbf z_i)+\log r_i^{*(k-1)}.
\)

\item {KAN fitting:}
\(
\hat\eta^{(k)}
=
\arg\min_{\eta}
\frac{1}{n}\sum_{i=1}^n
\left(Y_i^{*(k)}-\mathrm{KAN}(\mathbf z_i)\right)^2 .
\)

\item {Repeat} until
\(
\frac1n\sum_{i=1}^n
\left|\hat\eta^{(k)}(\mathbf z_i)-\hat\eta^{(k-1)}(\mathbf z_i)\right|
<\tau ,
\)
\end{enumerate}
where $\tau>0$ denotes a convergence tolerance, set to $10^{-3}$ in our implementation. This procedure yields $\hat\eta(\mathbf z)$, the Buckley--James KAN-AFT estimator.

%%%%%%%%%%%%%%%%%%%%%%%%%%%%%%%%%%%%%%%%%%%%%%%%%%%%%%%%%%%%%%
\subsection{KAN-AFT--IPCW}

The IPCW (Inverse Probability of Censoring Weighting) method corrects for censoring by weighting uncensored observations by the
inverse probability of being uncensored \cite{ipcw}.

Let $\hat G(t)$ denote the Kaplan--Meier estimator of the censoring survival function
$G(t)=P(C>t)$. Define weights
\(
w_i=\frac{\delta_i}{\hat G(\tilde T_i)}.
\)
The weighted loss is given by
$$L_{\mathrm{IPCW}}
=
\frac{1}{n}\sum_{i=1}^n
w_i\left(Y_i-\mathrm{KAN}(\mathbf z_i)\right)^2.
$$
Only uncensored observations contribute, while the weights correct selection bias.
This estimator is non-iterative and computationally efficient.

%%%%%%%%%%%%%%%%%%%%%%%%%%%%%%%%%%%%%%%%%%%%%%%%%%%%%%%%%%%%%%
\subsection{KAN-AFT--Transform (Transformation Method)}

Following Fan and Gijbels \cite{FanGijbels1996}, a transformed outcome is constructed
so that standard least-squares regression remains valid under censoring.

Let
\(
T_i^*=\phi_1(\tilde T_i)\delta_i+\phi_2(\tilde T_i)(1-\delta_i),
\)
where
\[
\phi_1(t)
=(1+\alpha)\int_0^t \frac{du}{\hat G(u)}-\alpha\frac{t}{\hat G(t)},
\qquad
\phi_2(t)
=(1+\alpha)\int_0^t \frac{du}{\hat G(u)} .
\]

The constant $\alpha$ is chosen so that $T_i^*\ge0$ for all $i$:
\[
\alpha=
\min_{i:\delta_i=1}
\frac{
\int_0^{\tilde T_i}\hat G(u)^{-1}du-\tilde T_i
}{
\tilde T_i\hat G(\tilde T_i)^{-1}
-\int_0^{\tilde T_i}\hat G(u)^{-1}du
}.
\]

The KAN model is trained using
\(
L_{\mathrm{Trans}}
=
\frac{1}{n}\sum_{i=1}^n
\left(\log T_i^*-\mathrm{KAN}(\mathbf z_i)\right)^2 .
\)

This approach avoids iterative imputation and weighting while retaining consistency.

\section{Hyperparameter Specifications}\label{append_3}

This section summarizes the hyperparameter settings adopted for the KAN--AFT models across all real-data applications. Dimensions, denote the structural tuple $(n_{\text{cov}}, n_h, 1)$, corresponding to the numbers of covariate components, latent aggregation components, and the output regression component, where the case $n_h=0$ yields a purely additive representation. Spline degree (d) specifies the degree of the spline basis, grid size (G) the number of spline knots per component, decay ($\lambda_2$) the regularization parameter, and patience specifies the number of epochs with no improvement permitted before early stopping is applied.
%Latent aggregation components are composite indices formed by aggregating transformed covariates and mapping each aggregate through a smooth univariate function.

\begin{table}[!thpb]
\centering
\caption{Hyperparameter settings for KAN--AFT models across real-data applications}
\begin{tabular}{l l c c c c c}
\toprule
Dataset & Censoring method & Dimensions & d & G & $\lambda_2$ & Patience \\
\midrule
\multirow{3}{*}{Veteran}
& Buckley James & (3, 0, 1) & 3 & 15 & 0.06 & 5\\
& IPCW & (3, 0, 1) & 3 & 5 & 0.05 & 5 \\
& Transform & (3, 0, 1) & 5 & 15 & 0.05 & 5 \\
\hline
\multirow{3}{*}{GBSG}
& Buckley James & (6, 0, 1) & 3 & 5 & 0.05 & 15 \\
& IPCW & (6, 0, 1) & 3 & 5 & 0.05 & 15 \\
& Transform & (6, 0, 1) & 3 & 5 & 0.04 & 15\\
\hline
\multirow{3}{*}{PBC}
& Buckley James & (10, 3, 1) & 3 & 15 & 0.05 & 5 \\
& IPCW & (10, 3, 1) & 3 & 15 & 0.02 & 5 \\
& Transform & (10, 2, 1) & 3 & 15 & 0.05 & 5\\
\hline
\multirow{3}{*}{Heart Failure}
& Buckley James & (6, 2, 1) & 3 & 15 & 0.05 & 7 \\
& IPCW & (6, 3, 1) & 3 & 15 & 0.02 & 5 \\
& Transform & (6, 2, 1) & 3 & 15 & 0.03 & 5\\
\bottomrule
\end{tabular}
\label{tab:hyperparameters_all}
\end{table}

\newpage
\bibliography{Rererece}
%\newpage
%%%%%%%%%%%%%%%%%%%%%%%%%%%%%%%%%%%%%%%%%%%%%%%%%%%%%%%%%%%%%%%%%%%%%%%%%%%%%%%%%
%%%%%%%%%%%%%%%%%%%%%%%%%%%%%%%%%%%%%%%%%%%%%%%%%%%%%%%%%%%%%%%%%%%%%%%%%%%%%%%%%

%%%%%%%%%%%%%%%%%%%%%%%%%%%%%%%%%%%%%%%%%%%%%%%%%%%%%%

%%%%%%%%%%%%%%%%%%%%%%%%%%%%%%%%%%%%%%%%%%%%%%%%%%%%%%

%%%%%%%%%%%%%%%%%%%%%%%%%%%%%%%%%%%%%%%

\end{document}